\PassOptionsToPackage{table}{xcolor}
\documentclass{article}
\usepackage[T1]{fontenc}
\usepackage{iclr2027_conference,times}
\usepackage{amsmath,amssymb,booktabs,array,graphicx,multirow,tabularx,capt-of,wrapfig}
\usepackage{algorithm}
\usepackage{algpseudocode}
\usepackage{xcolor}
\usepackage{tikz}
\usepackage{hyperref,url}
\definecolor{rowgray}{HTML}{DFDFDF}
\definecolor{pulseaccent}{HTML}{A31F34}
\definecolor{appendixred}{HTML}{D62728}
\newcommand{\W}{\mathbf{W}}
\newcommand{\X}{\mathbf{X}}
\newcommand{\x}{\mathbf{x}}
\newcommand{\D}{\mathbf{D}}
\newcommand{\Rot}{\boldsymbol{\Pi}_d}
\newcommand{\method}{PulseQuant}

\newcommand{\best}[1]{\textbf{#1}}
\newcommand{\second}[1]{\underline{#1}}
\newcommand{\R}{\mathbb{R}}
\DeclareRobustCommand{\discretepulsemark}{%
  \begin{tikzpicture}[x=0.032em,y=-0.032em,baseline=-0.80em]
    \fill[pulseaccent] (14,14) circle[radius=14];
    \fill[white,rounded corners=0.025em] (5.5,12) rectangle (7.5,16);
    \fill[white,rounded corners=0.025em] (9.25,9) rectangle (11.25,19);
    \fill[white,rounded corners=0.025em] (13,6) rectangle (15,22);
    \fill[white,rounded corners=0.025em] (16.75,9) rectangle (18.75,19);
    \fill[white,rounded corners=0.025em] (20.5,12) rectangle (22.5,16);
  \end{tikzpicture}%
}
\title{\discretepulsemark\hspace{0.25em}PulseQuant: Propagation-Guided Subspace Correction for 4-Bit Video Diffusion Transformers}

\author{Yutong Wang\textsuperscript{1}\quad 
Xingtong Ge\textsuperscript{2}\quad 
Enhuai Liu\textsuperscript{1}\quad 
Yunke Wang\textsuperscript{1}\quad 
Tianfan Xue\textsuperscript{4,3}\\ 
\textbf{Xinyuan Chen\textsuperscript{3}\footnotemark[1] \quad 
Chang Xu\textsuperscript{1}\thanks{Corresponding authors.}}\\
\textsuperscript{1}The University of Sydney\quad 
\textsuperscript{2}The Hong Kong University of Science and Technology \\ 
\textsuperscript{3}Shanghai AI Laboratory\quad 
\textsuperscript{4}The Chinese University of Hong Kong
}

\iclrfinalcopy 
\begin{document}
\maketitle

\begin{abstract}
Quantization errors in video diffusion transformers can be amplified or attenuated by subsequent denoising updates, making local reconstruction error an incomplete predictor of final impact. We introduce \method{}, a 4-bit post-training quantization method that combines trajectory sensitivity with activation geometry to guide offline calibration. Isolated block--step interventions estimate propagation risk, which prioritizes sensitive trajectory states during row-radius selection. With these radii fixed, response-subspace correction uses neighboring-code edits to reduce residual components along dominant activation directions. Both stages preserve the original 4-bit weight representation. Controlled interventions show that short-horizon propagated error predicts final latent error more reliably than immediate block-output error, supporting calibration beyond local reconstruction objectives. Evaluations on Wan models, Self Forcing, and MiniMax-H3 demonstrate improvements in key consistency and dense-reference metrics while remaining competitive on other attributes across model scales and generation paradigms.
Code and models are available at \url{https://github.com/hhhh1138/PulseQuant}.
\end{abstract}


\section{Introduction}
Video diffusion transformers~\citep{peebles2023dit,wan2025,huang2026self,minimax2026h3} repeatedly execute a large network throughout denoising, making post-training quantization (PTQ) attractive for reducing storage and computation without retraining. Yet the same iterative process can amplify quantization errors: a small local disturbance may develop into spatial distortion or temporal inconsistency after subsequent updates. Video PTQ must therefore preserve the denoising trajectory, not just the accuracy of isolated operators.

Existing methods adapt to timestep-varying distributions, reshape tensors before quantization, or preserve difficult structure in higher precision~\citep{li2023qdiffusion,he2023ptqd,zhao2024viditq,xiao2023smoothquant,li2024svdquant}. These strategies improve local quantization behavior, but local reconstruction quality alone does not establish how an error will affect the remaining denoising trajectory. Subsequent updates may attenuate or amplify similar local disturbances differently, creating a \emph{\textbf{local-to-global misalignment}} between immediate error and final impact.

\begin{figure*}[t]
\centering
\includegraphics[width=\textwidth]{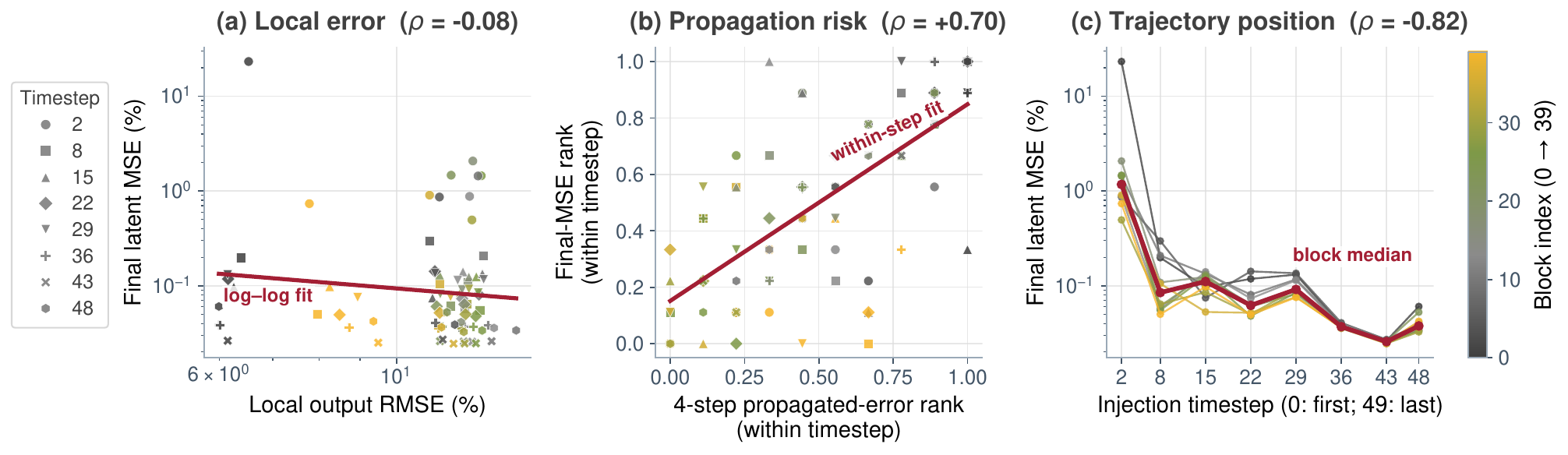}
\caption{Propagation diagnostics from 80 isolated 4-bit interventions on Wan~2.1-14B (10 blocks $\times$ eight denoising steps), each followed by dense continuation. (a) Local block-output RMSE is weakly associated with final latent error. (b) Four-step propagated error $E_H$ predicts final damage after ranking blocks within each timestep. (c) Final damage depends strongly on trajectory position. Colors encode block depth and markers encode injection timestep.}
\label{fig:propagation-evidence}
\end{figure*}

To examine this mismatch, we perform 80 isolated 4-bit block--step interventions on Wan~2.1-14B, each followed by dense continuation. Figure~\ref{fig:propagation-evidence} shows that local block-output RMSE is nearly uncorrelated with final latent error ($\rho=-0.08$), whereas four-step propagated error better recovers its within-timestep ranking ($\rho=0.70$); final error also varies strongly with trajectory position. This evidence motivates weighting calibration by downstream sensitivity. Within the fixed 4-bit representation, calibration must then choose code changes that effectively reduce response error, which depends on how weight residuals align with the activation distribution.

We introduce \method, a propagation-guided subspace correction method for 4-bit video diffusion transformers. As illustrated in Figure~\ref{fig:trajectory-motivation}, \emph{\textbf{propagation-aware calibration}} uses isolated block--step interventions to prioritize sensitive trajectory states during row-radius selection. With these radii fixed, \emph{\textbf{response-subspace correction}} applies legal neighboring-code edits to reduce residual components along dominant activation directions. Both stages operate offline and preserve the existing 4-bit weight representation, without high-precision residual branches or online correction.
Across model scales, generation paradigms, and activation precisions, \method{} improves key consistency and dense-reference metrics, leading all paired fidelity metrics at W4A4 among the tested PTQ baselines. For Wan~2.1-1.3B on an RTX~5080, packed NVFP4 more than halves DiT peak memory; combining it with low-bit attention~\citep{li2026vcattention,zhang2024sageattention2} yields up to a 3.51$\times$ DiT-step speedup over Dense BF16.

Our contributions are threefold:
\begin{itemize}
    \item We quantify the local-to-global misalignment in video PTQ through isolated 4-bit block--step interventions. Short-horizon propagated error predicts final latent error more reliably than immediate reconstruction error, motivating calibration that accounts for downstream sensitivity across trajectory positions.
    \item We develop an offline calibration procedure that couples propagation-weighted radius selection with response-subspace code correction. The former prioritizes sensitive trajectory states, while the latter uses neighboring-code edits to reduce response error along dominant activation directions at fixed radii. 
    \item We demonstrate gains in key consistency and dense-reference metrics across bidirectional and autoregressive video generators, model scales, and activation precisions. Native packed NVFP4 deployment on an RTX~5080 more than halves DiT peak memory and complements low-bit attention, delivering up to a 3.51$\times$ DiT-step speedup over Dense BF16.
\end{itemize}

\section{Related Work}
\subsection{Quantization for diffusion and video transformers}
Diffusion PTQ must accommodate activation distributions that change throughout denoising. Diffusion-specific calibration and error correction address this nonstationarity~\citep{shang2023ptq4dm,li2023qdiffusion,he2023ptqd}, while timestep-aware quantization and reconstruction explicitly account for temporal variation~\citep{so2023tdq,huang2024tfmq,wang2024apqdm}. For image and video transformers, this challenge extends to variation across spatial positions, channels, and tokens, motivating finer-grained quantization and calibration strategies~\citep{chen2025qdit,zhao2024viditq,feng2025qvdit,li2025dvdquant}. Spatiotemporal redundancy offers another compression opportunity, exploited by DeltaQuant~\citep{li2026deltaquant}. Beyond PTQ, QVGen~\citep{huang2026qvgen} uses quantization-aware training with auxiliary correction modules that are progressively removed through rank decay. Alongside improving the low-bit approximation, calibration must account for how its errors evolve during iterative generation.

Error accumulation across denoising steps is an established concern~\citep{li2023qdiffusion}. AccuQuant~\citep{lee2025accuquant} addresses it directly by matching full-precision and quantized outputs over multiple simulated denoising steps. \method{} instead uses isolated block--timestep pulses followed by dense continuation to estimate downstream sensitivity and incorporates these measurements as weights in row-radius calibration.

\subsection{Low-bit representations and response correction}
Low-bit PTQ improves quantization quality by reshaping the representation or optimizing discrete reconstruction decisions. Channel scaling and learnable equivalent transformations redistribute quantization difficulty while preserving the dense function~\citep{xiao2023smoothquant,shao2024omniquant}, with AWQ~\citep{lin2024awq} using activation-informed scaling to protect salient weight channels. Fixed or learned rotations further reduce outlier concentration~\citep{ashkboos2024quarot,liu2024spinquant}. Incoherence processing and structured codebooks support accurate low-bit representations~\citep{tseng2024quipsharp}, while AQLM~\citep{egiazarian2024aqlm} learns additive codebooks with input-adaptive reconstruction and block-wise joint optimization. For visual diffusion transformers, OrbitQuant combines normalization, RPBH rotation, and a shared spherical codebook~\citep{lee2026orbitquant}.

Within a chosen representation, reconstruction-based methods optimize rounding or code decisions using weight-, layer-, or block-local objectives~\citep{nagel2020adaround,li2021brecq,frantar2022gptq}. SVDQuant instead retains difficult structure in a high-precision low-rank branch~\citep{li2024svdquant}. Building on the OrbitQuant representation, \method{} uses activation-adapted response directions to guide neighboring-code edits at fixed calibrated row radii. The subspace guides offline index selection, while the exported weights retain the shared codebook and per-row scales.

\section{Preliminaries}
\label{sec:prelim}
\subsection{Quantized linear projections}
Consider a linear layer $\mathbf y=\W\x$, where $\x\in\R^d$, $\mathbf y\in\R^m$, and $\W\in\R^{m\times d}$. We adopt the channel transform of OrbitQuant~\citep{lee2026orbitquant}: a positive diagonal scaling $\D$ followed by an orthogonal randomized permuted block-Hadamard (RPBH) rotation $\Rot$. Transforming the activations and applying the corresponding inverse transform to the weights gives

\begin{equation}
 \x'=\Rot\D\x,\qquad
 \W'=\W\D^{-1}\Rot^\top,\qquad
 \W'\x'=\W\x.
 \label{eq:transform}
\end{equation}
These transforms reshape the quantization coordinates while preserving the dense layer output. For a token batch $\X\in\R^{N\times d}$ stored row-wise, the activation transform is $\X'=\X\D\Rot^\top$. The low-bit layer computes
\begin{equation}
 \widehat{\mathbf y}=\widehat{\W}'Q_A(\x'),
 \label{eq:inference}
\end{equation}
where $Q_A$ denotes symmetric per-token activation quantization, expressed in dequantized values, and $\widehat\W'$ denotes the quantized weight matrix represented by 4-bit indices and per-row scales. 
Appendix~\ref{app:protocol} specifies the activation quantizer.

\subsection{Spherical weight coding}
OrbitQuant~\citep{lee2026orbitquant} normalizes each transformed weight row $\mathbf w'_i$ by its initial radius $r_i=\|\mathbf w'_i\|_2$. Each normalized coordinate is then assigned to its nearest entry in a shared, ordered 16-value codebook $\mathcal C=\{C_0,\ldots,C_{15}\}$:
\begin{equation}
 c_{ij}=\arg\min_{k\in\{0,\ldots,15\}}|w'_{ij}/r_i-C_k|,
 \qquad q_{ij}=r_iC_{c_{ij}}.
 \label{eq:spherical-code}
\end{equation}
Each row is therefore stored as one radius, serving as its reconstruction scale, and one 4-bit index per coordinate. \method{} calibrates the row radii and refines the discrete indices while keeping the codebook fixed; the resulting weights are $\widehat w'_{ij}=\bar r_iC_{c_{ij}}$.

\section{Method}
\label{sec:method}
\subsection{Overview}
\begin{figure*}[t]
    \centering
    \includegraphics[width=\textwidth]{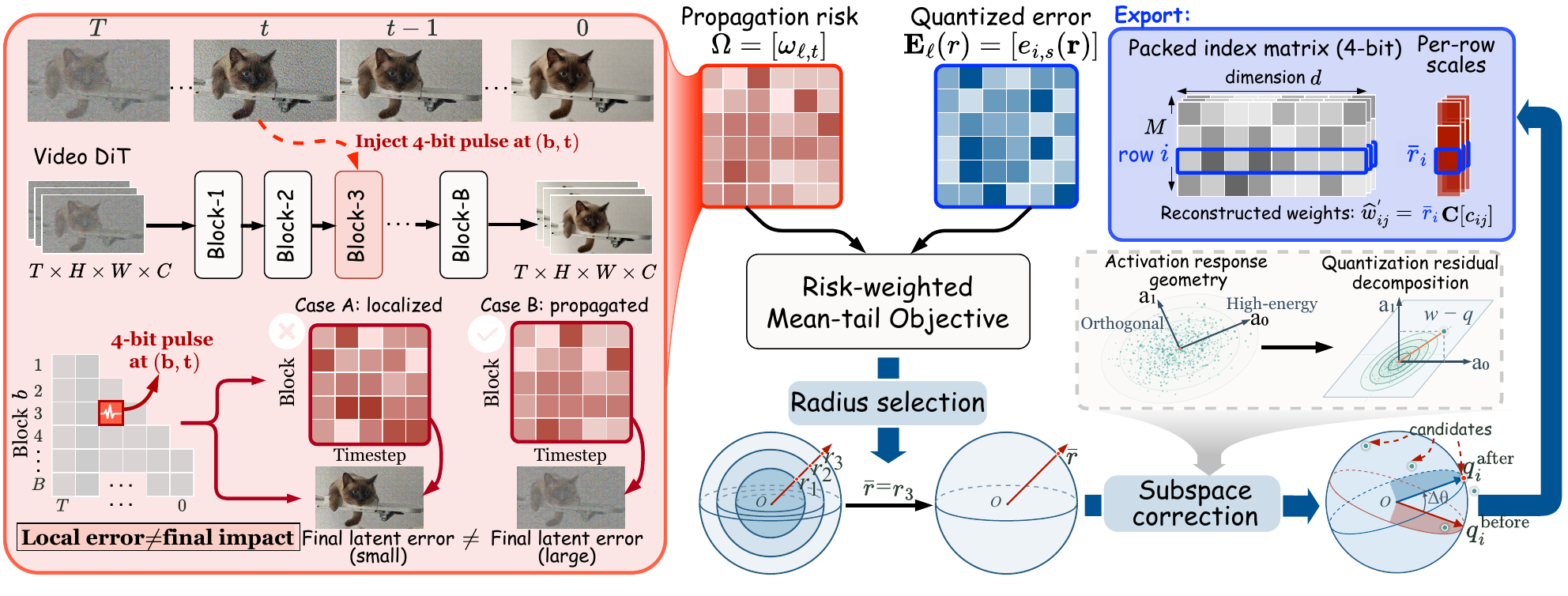}
    \caption{Overview of \method{}. Isolated 4-bit pulses reveal how local perturbations propagate through denoising (left). The resulting risk weights guide per-row radius selection through a mean--tail output-error objective (center). With the selected radii fixed, legal neighboring-code edits reduce residual components along high-energy activation directions (right). The calibrated weights are exported as packed 4-bit indices and per-row scales, with no online profiling or correction.}
    \label{fig:trajectory-motivation}
\end{figure*}

\method{} calibrates row radii and discrete weight indices in two offline stages, as illustrated in Figure~\ref{fig:trajectory-motivation}. \emph{Propagation-aware calibration} first profiles isolated 4-bit perturbations and uses their gains to weight row-radius selection. \emph{Response-subspace correction} then holds the selected radii fixed and applies legal neighboring-code edits to reduce residual components along dominant activation directions. The resulting indices and row radii define the quantized weights used in Eq.~\ref{eq:inference}. Algorithm~\ref{alg:pipeline} gives the complete procedure; Figure~\ref{fig:trajectory-diagnostics} in Appendix~\ref{app:component-full} visualizes activation balancing, propagation risk, and residual reduction.

\subsection{Propagation-aware calibration}
\label{sec:propagation-risk}
\paragraph{Profiling trajectory sensitivity.}
We estimate downstream sensitivity with an isolated intervention at each sampled block--timestep anchor $(b,t)$. Starting from the same latent state, we compare a dense trajectory with a pulsed trajectory that quantizes block $b$ only at step $t$; both then continue in dense precision for $H$ updates. The profiling pulse uses uniform W4A4 quantization--dequantization on the attention and FFN linear projections of the selected block. For CFG-based profiling, it covers both forward calls at the anchor step; Appendix~\ref{app:profiling-pulse} details the pulse configuration. Let $\mathbf z^{\mathrm p}$ and $\mathbf z^{\mathrm d}$ denote their latent states. We measure the normalized discrepancy immediately after the pulse and after $H$ further updates, and define the propagation gain from their relative change:
\begin{equation}
 E_h(b,t)=\frac{\|\mathbf z^{\mathrm p}_{t+1+h}-\mathbf z^{\mathrm d}_{t+1+h}\|_2^2}
 {\|\mathbf z^{\mathrm d}_{t+1+h}\|_2^2+\epsilon},\quad h\in\{0,H\},
 \qquad g_{b,t}=\sqrt{\frac{E_H(b,t)+\epsilon}{E_0(b,t)+\epsilon}}.
\label{eq:gain}
\end{equation}
The gain measures the change in normalized latent discrepancy over $H$ subsequent updates: $g_{b,t}>1$ indicates amplification, whereas $g_{b,t}<1$ indicates attenuation.
We obtain a trajectory-wide profile by interpolating positive gains from a sparse anchor grid in log space over the scheduler noise coordinate and relative block depth. Each layer $\ell$ inherits the interpolated profile $\bar g_{\ell,t}$ of its block.
We temper and clip the profile, then normalize the weights to have unit mean over all block--timestep locations:
\begin{equation}
 \omega_{\ell,t}=\operatorname{MeanOne}\!\left(
 \operatorname{clip}(\bar g_{\ell,t}^{\gamma},a,b)\right).
 \label{eq:weights}
\end{equation}
Here, $\gamma$ controls the contrast between locations, $[a,b]$ limits the influence of extreme estimates, and mean-one normalization preserves the overall objective scale. The resulting $\omega_{\ell,t}$ assigns greater calibration weight to perturbations that are more strongly amplified downstream. Appendix~\ref{app:derivations} relates this finite-horizon measurement to first-order error propagation.

\paragraph{Risk-weighted radius selection.}
The propagation profile weights the contribution of each trajectory state to radius calibration. For layer $\ell$, let $\X'_{\ell,s}\in\R^{N_s\times d}$ contain the transformed tokens recorded at calibration call $s$, and let $t(s)$ be the corresponding denoising step. For row $i$, the base indices define the fixed codebook direction $\mathbf v_i=[C_{c_{i1}},\ldots,C_{c_{id}}]^\top$, giving the reconstruction $r\mathbf v_i$ at radius $r$. For each candidate $r\in\mathcal R_i$, we compute the mean squared row-output error $e_{i,s}(r)$ and weight it by the propagation risk at that call:
\begin{equation}
\begin{aligned}
 e_{i,s}(r)&=\frac{1}{N_s}\|\X'_{\ell,s}(\mathbf w'_i-r\mathbf v_i)\|_2^2, \qquad
 u_{i,s}(r)=\omega_{\ell,t(s)}e_{i,s}(r),\\
 \bar r_i&=\arg\min_{r\in\mathcal R_i}
 \left[(1-\lambda)\operatorname{Mean}_s u_{i,s}(r)
 +\lambda\operatorname{TopMean}_{\rho,s}u_{i,s}(r)\right].
\end{aligned}
\label{eq:radius-objective}
\end{equation}
Here, $\operatorname{TopMean}_{\rho}$ averages the largest $\rho$ fraction of the risk-weighted call errors, and $\lambda$ balances the mean and tail terms. The mean term controls average response distortion, while the tail term emphasizes calls with the largest risk-weighted errors. Propagation weighting and tail aggregation serve distinct roles: the former accounts for downstream sensitivity, while the latter prioritizes calls that remain difficult after this weighting. Because each weight row is reused throughout denoising, the objective selects a single radius across the sampled trajectory states. We hold this radius $\bar r_i$ fixed during the subsequent code correction.

\subsection{Response-subspace correction}
\label{sec:pca}
With the row radii calibrated, we refine the discrete indices using the activation geometry. Weight residuals affect the layer output through their interaction with activations: a residual aligned with a high-energy activation direction produces a larger mean squared output error than an equally sized residual along a low-energy direction. This motivates focusing code correction on a compact set of high-energy response directions. For layer $\ell$, we retain an evenly spaced subset $\bar\X'_\ell\in\R^{M_\ell\times d}$ and define its uncentered second moment $\mathbf C_\ell=\bar\X_\ell^{\prime\top}\bar\X'_\ell/M_\ell$. The quantity $\mathbf a^\top\mathbf C_\ell\mathbf a$ measures activation energy along direction $\mathbf a$. We estimate two high-energy directions with a short orthogonal power iteration, initialized by the all-ones and alternating-sign Walsh patterns:
\begin{equation}
\begin{aligned}
 \mathbf a_0&\leftarrow\operatorname{norm}(\mathbf C_\ell\mathbf a_0),
 &\mathbf a_0^{(0)}&\propto[1,1,\ldots],\\
 \mathbf a_1&\leftarrow\operatorname{norm}\!\left(
 (\mathbf I-\mathbf a_0\mathbf a_0^\top)\mathbf C_\ell\mathbf a_1\right),
 &\mathbf a_1^{(0)}&\propto[1,-1,\ldots].
\end{aligned}
\label{eq:pca-iteration}
\end{equation}
Multiplication by $\mathbf C_\ell$ adapts each seed toward a high-energy activation direction, while the second update orthogonalizes the second axis against the first. We use $M_{\max}=512$ tokens and two iterations, evaluating each matrix product as $\bar\X_\ell^{\prime\top}(\bar\X'_\ell\mathbf a)/M_\ell$ without materializing $\mathbf C_\ell$.

The resulting axes guide discrete correction at the selected radius $\bar r_i$. We partition row $i$ into channel groups $\mathcal G$ of size $G$; within a group, let $\mathbf w$ and $\mathbf q$ denote the corresponding slices of the dense transformed row $\mathbf w'_i$ and its current quantized reconstruction, and restrict each $\mathbf a_u$ in the same way. We minimize
\begin{equation}
 \mathcal L_{i,\mathcal G}(\mathbf q)=
 \underbrace{
 \frac{1}{2}\sum_{u=0}^{1}
 \frac{\langle\mathbf w-\mathbf q,\mathbf a_u\rangle^2}
 {\|\mathbf a_u\|_2^2+\epsilon}
 }_{\text{subspace response}}
 +\underbrace{
 \frac{\tau}{G}\|\mathbf w-\mathbf q\|_2^2
 }_{\text{full-group distortion}}.
 \label{eq:joint-correction}
\end{equation}
Here, $\mathbf w-\mathbf q$ is the group-wise residual between the dense transformed row and its current 4-bit reconstruction. The first term averages its squared projections onto the two response axes, normalized by their squared norms with $\epsilon$ for stability. The second term controls mean squared distortion across the full group, with weight $\tau$. Each legal neighboring-code edit replaces an index with an adjacent codebook index at the fixed row scale. From the same current reconstruction, we evaluate both one-level moves for each coordinate and retain the better proposal. We rank these proposals by objective decrease and select the cumulative prefix with the lowest joint objective, including the empty prefix. Each prefix is evaluated jointly to account for interactions among coordinate edits. Appendix~\ref{app:default-hyperparameters} reports the default settings, while Appendix~\ref{app:discrete-search} provides the exact updates, response-error bound, and prefix-selection guarantee.

\medskip
\noindent
\begin{minipage}[t]{0.43\textwidth}
\vspace{0pt}
\subsection{Calibration and deployment}
Algorithm~\ref{alg:pipeline} summarizes offline calibration. Recorded activations support both radius selection and response-axis estimation, while the propagation profile weights calibration calls by downstream sensitivity. With $K_{\mathrm p}$ power iterations and $M_\ell$ retained tokens, axis estimation costs $O(2K_{\mathrm p}M_\ell d)$ per layer; correction for an $m\times d$ weight matrix costs $O(md\log G)$ per pass. These costs are incurred offline. The calibrated indices and row radii, together with the transform metadata, define the exported model used in Eq.~\ref{eq:inference}. Each layer shares its calibrated weights across denoising steps, while propagation profiles and response axes guide offline calibration.
\end{minipage}\hfill%
\begin{minipage}[t]{0.53\textwidth}
\vspace{0pt}
\hrule\vspace{3pt}
\refstepcounter{algorithm}\label{alg:pipeline}
\textbf{Algorithm \thealgorithm}\quad \method{} offline calibration
\vspace{3pt}\hrule\vspace{2pt}
\small
\begin{algorithmic}[1]
\Require Model $f$, calibration set $\mathcal D$, transforms $(\D,\Rot)$, codebook $\mathcal C$, anchors $\mathcal A$
\Ensure Corrected indices $\{c_{ij}\}$, row radii $\{\bar r_i\}$ 
\State Run dense $f$ on $\mathcal D$; record activations $\X'_{\ell,s}$
\For{each block--step anchor $(b,t)\in\mathcal A$}
    \State Apply one 4-bit pulse; compute $g_{b,t}$ by Eq.~\ref{eq:gain}
\EndFor
\State Interpolate and normalize $\{g_{b,t}\}$ into $\omega_{\ell,t}$ by Eq.~\ref{eq:weights}
\For{each target layer $\ell$}
    \State Initialize $\{c_{ij}\}$ by Eq.~\ref{eq:spherical-code}; select $\{\bar r_i\}$ by Eq.~\ref{eq:radius-objective}
    \State Estimate $(\mathbf a_{\ell,0},\mathbf a_{\ell,1})$ by Eq.~\ref{eq:pca-iteration}
    \For{each row $i$ and channel group $\mathcal G$}
        \State Rank code moves by decrease in Eq.~\ref{eq:joint-correction}
        \State Apply the minimum-loss cumulative prefix
    \EndFor
\EndFor
\State \Return $\{c_{ij}\}$, $\{\bar r_i\}$, and transform metadata
\end{algorithmic}
\vspace{2pt}\hrule
\end{minipage}

\begin{table*}[t]
\centering
\caption{\textbf{Quantitative evaluations on VBench (T2V).} VBench attributes are percentages; PSNR, SSIM, and LPIPS compare matched quantized and Dense BF16 videos. We bold the \textbf{best results} and underline the \underline{second-best results}.}
\label{tab:vbench}
\setlength{\tabcolsep}{2.2pt}
\renewcommand{\arraystretch}{1.06}
\resizebox{0.92\textwidth}{!}{%
\begin{tabular}{llc@{\hspace{5pt}}ccccccc@{\hspace{7pt}}ccc}
\toprule
\multirow[c]{2}{*}{Model} & \multirow[c]{2}{*}{Method} & \multirow[c]{2}{*}{W/A} & \multicolumn{7}{c}{VBench attributes (\%)} & \multicolumn{3}{c}{Dense-reference fidelity} \\
\cmidrule(lr){4-10}\cmidrule(lr){11-13}
 & & & IQ $\uparrow$ & AQ $\uparrow$ & MS $\uparrow$ & BC $\uparrow$ & SC $\uparrow$ & Scene $\uparrow$ & OC $\uparrow$ & PSNR $\uparrow$ & SSIM $\uparrow$ & LPIPS $\downarrow$ \\
\midrule
\multirow{13}{*}{Wan~2.1-1.3B}
 & Dense BF16 & 16/16 & 67.71 & 66.06 & 98.29 & 96.19 & 93.97 & 42.44 & 26.15 & ref. & ref. & ref. \\
\cmidrule(lr){2-13}
 & SmoothQuant~\citep{xiao2023smoothquant} & 4/6 & 59.83 & 58.12 & 98.12 & 94.77 & 91.14 & 30.96 & 24.20 & 14.78 & 0.464 & 0.499 \\
 & SVDQuant~\citep{li2024svdquant} & 4/6 & 63.68 & 63.57 & 97.83 & 95.20 & 92.62 & 35.61 & 25.42 & \second{15.36} & 0.513 & 0.433 \\
 & ViDiT-Q~\citep{zhao2024viditq} & 4/6 & 63.60 & 63.68 & 97.78 & 95.03 & 92.70 & 38.08 & \second{25.73} & 14.86 & 0.503 & 0.447 \\
 & DVD-Quant~\citep{li2025dvdquant} & 4/6 & 63.21 & 64.04 & \best{98.62} & 95.23 & \second{94.38} & 41.86 & 25.60 & 14.96 & \best{0.529} & 0.430 \\
 & OrbitQuant~\citep{lee2026orbitquant} & 4/6 & \second{63.78} & \second{64.72} & \second{98.54} & \second{95.37} & \best{94.66} & \second{42.01} & 25.57 & 15.30 & 0.521 & \second{0.424} \\
 & \cellcolor{rowgray}\method\ (ours) & \cellcolor{rowgray}4/6 & \cellcolor{rowgray}\best{64.96} & \cellcolor{rowgray}\best{64.79} & \cellcolor{rowgray}98.21 & \cellcolor{rowgray}\best{95.44} & \cellcolor{rowgray}93.10 & \cellcolor{rowgray}\best{43.97} & \cellcolor{rowgray}\best{25.91} & \cellcolor{rowgray}\best{15.64} & \cellcolor{rowgray}\second{0.527} & \cellcolor{rowgray}\best{0.411} \\
\cmidrule(lr){2-13}
 & SmoothQuant~\citep{xiao2023smoothquant} & 4/4 & 55.24 & 47.60 & 94.35 & 94.56 & 85.76 & 11.77 & 19.76 & 14.38 & 0.392 & 0.561 \\
 & SVDQuant~\citep{li2024svdquant} & 4/4 & 62.36 & 59.61 & 96.94 & \second{94.76} & 90.19 & 20.09 & 24.32 & \second{15.27} & 0.479 & 0.465 \\
& ViDiT-Q~\citep{zhao2024viditq} & 4/4 & 58.02 & 50.23 & 95.23 & 94.59 & 86.03 & 15.19 & 20.99 & 14.30 & 0.401 & 0.547 \\
 & DVD-Quant~\citep{li2025dvdquant} & 4/4 & 51.18 & 52.74 & \best{98.28} & 94.24 & 89.88 & 13.30 & 21.66 & 12.96 & 0.392 & 0.658 \\
 & OrbitQuant~\citep{lee2026orbitquant} & 4/4 & \best{63.14} & \best{63.33} & \second{98.10} & 94.69 & \best{93.76} & \second{34.23} & \second{25.17} & 14.97 & \second{0.491} & \second{0.458} \\
 & \cellcolor{rowgray}\method\ (ours) & \cellcolor{rowgray}4/4 & \cellcolor{rowgray}\second{62.70} & \cellcolor{rowgray}\second{62.46} & \cellcolor{rowgray}97.83 & \cellcolor{rowgray}\best{94.84} & \cellcolor{rowgray}\second{91.97} & \cellcolor{rowgray}\best{36.70} & \cellcolor{rowgray}\best{25.44} & \cellcolor{rowgray}\best{15.42} & \cellcolor{rowgray}\best{0.503} & \cellcolor{rowgray}\best{0.443} \\

\midrule

\multirow{13}{*}{Self Forcing}
 & Dense BF16 & 16/16 & 70.88 & 67.33 & 98.40 & 95.53 & 94.77 & 47.31 & 26.42 & ref. & ref. & ref. \\
\cmidrule(lr){2-13}
 & SmoothQuant~\citep{xiao2023smoothquant} & 4/6 & 70.40 & 66.15 & 98.12 & 94.80 & 94.01 & 46.29 & 26.21 & 13.71 & 0.402 & 0.464 \\
 & SVDQuant~\citep{li2024svdquant} & 4/6 & 70.33 & 66.05 & 98.50 & 95.33 & 95.04 & \best{50.07} & 26.17 & 13.81 & 0.401 & 0.469 \\
  & ViDiT-Q~\citep{zhao2024viditq} & 4/6 & 70.40 & 65.97 & 97.94 & 94.91 & 93.60 & 43.53 & 26.20 & 13.51 & 0.392 & 0.495 \\
 & DVD-Quant~\citep{li2025dvdquant} & 4/6 & \best{70.64} & 66.02 & \second{98.52} & \second{95.60} & \best{95.37} & 47.97 & 26.06 & 13.69 & \second{0.409} & 0.457 \\
 & OrbitQuant~\citep{lee2026orbitquant} & 4/6 & 70.08 & \best{67.02} & 98.40 & 95.51 & 94.57 & \second{49.49} & \second{26.26} & \second{14.02} & 0.408 & \second{0.454} \\
 & \cellcolor{rowgray}\method\ (ours) & \cellcolor{rowgray}4/6 & \cellcolor{rowgray}\second{70.46} & \cellcolor{rowgray}\second{66.74} & \cellcolor{rowgray}\best{98.77} & \cellcolor{rowgray}\best{95.99} & \cellcolor{rowgray}\second{95.11} & \cellcolor{rowgray}45.64 & \cellcolor{rowgray}\best{26.28} & \cellcolor{rowgray}\best{14.17} & \cellcolor{rowgray}\best{0.428} & \cellcolor{rowgray}\best{0.443} \\
\cmidrule(lr){2-13}
 & SmoothQuant~\citep{xiao2023smoothquant} & 4/4 & 69.35 & 65.72 & 98.04 & 94.34 & 93.49 & 40.77 & 25.35 & 13.60 & 0.386 & 0.497 \\
 & SVDQuant~\citep{li2024svdquant} & 4/4 & 69.19 & 65.76 & 97.53 & 94.57 & 94.74 &\second{46.75}  & 25.83 & 13.80 & 0.392 & 0.495 \\
 & ViDiT-Q~\citep{zhao2024viditq} & 4/4 & 69.42 & 64.59 & 97.74 & 93.30 & 92.59 & 36.92 & 25.80 & 13.12 & 0.353 & 0.561 \\
 & DVD-Quant~\citep{li2025dvdquant} & 4/4 & \second{69.87} & 66.64 & \second{98.52} & 94.78 & \best{95.58} & 38.59 & 26.02 & 13.39 & 0.380 & 0.509 \\
 & OrbitQuant~\citep{lee2026orbitquant} & 4/4 & 69.04 & \best{67.02} & 98.51 & \best{95.18} & 94.77 & 44.69 & \second{26.20} & \second{13.91} & \second{0.401} & \second{0.476} \\
 & \cellcolor{rowgray}\method\ (ours) & \cellcolor{rowgray}4/4 & \cellcolor{rowgray}\best{69.96} & \cellcolor{rowgray}\second{66.80} & \cellcolor{rowgray}\best{98.85} & \cellcolor{rowgray}\second{95.10} & \cellcolor{rowgray}\second{95.32} & \cellcolor{rowgray}\best{47.60} & \cellcolor{rowgray}\best{26.33} & \cellcolor{rowgray}\best{14.07} & \cellcolor{rowgray}\best{0.418} & \cellcolor{rowgray}\best{0.461} \\
\bottomrule
\end{tabular}}

\end{table*}

\section{Experiments}
\label{sec:experiments}

\subsection{Implementation Details}
\paragraph{Models.}
We evaluate Wan~2.1-1.3B and 14B~\cite{wan2025}, Wan~2.2-A14B, Self Forcing~\cite{huang2026self}, and MiniMax-H3~\cite{minimax2026h3}, covering bidirectional and autoregressive generation across model scales and architectures.
We generate 81-frame videos for Wan models and Self Forcing, and 124-frame videos for MiniMax-H3. Wan models and MiniMax-H3 use classifier-free guidance (CFG), with default resolutions of $480\times832$ and $544\times960$, respectively.

\paragraph{Baselines and metrics.}
We compare \method{} with SmoothQuant~\citep{xiao2023smoothquant}, SVDQuant~\citep{li2024svdquant}, ViDiT-Q~\citep{zhao2024viditq}, DVD-Quant~\citep{li2025dvdquant}, and OrbitQuant~\citep{lee2026orbitquant}, using Dense BF16 as the reference.
We report seven VBench attributes~\citep{huang2024vbench} and paired PSNR, SSIM~\citep{wang2004ssim}, and LPIPS~\citep{zhang2018lpips} against Dense BF16 videos generated from the same noise.
VBench characterizes generated-video quality, while the paired metrics measure fidelity to the dense reference.

\paragraph{Precision and systems protocol.}
W4A4 and W4A6 use 4-bit weights and, respectively, 4- or 6-bit activations in every target linear layer. 
For all quality comparisons, we run every method on NVIDIA H200 using simulated quantization--dequantization (Q/DQ), with the same linear-layer coverage within each setting.
Within each setting, methods share evaluation prompts, seeds, and scheduler settings. Calibration statistics are re-estimated for each checkpoint and precision.
The systems evaluation is conducted on RTX~5080, where we deploy Wan~2.1-1.3B with packed NVFP4 weights~\citep{nvidia2025nvfp4} and native W4A4 Tensor Core kernels.
We report synchronized DiT-step latency and DiT-only peak allocation, excluding offline calibration and packing, text encoding, and VAE decoding. 
Section~\ref{sec:efficiency} reports the deployment results, while Appendix~\ref{app:deployment-details} provides kernel coverage and backend settings.

\subsection{Performance Analysis}

\noindent\textbf{Comparison with baselines.}
Table~\ref{tab:vbench} compares methods on Wan~2.1-1.3B and Self Forcing. 
In this table, we set $\lambda=0.75$ and $\rho=0.5$ as default. 
Across both models and precisions, \method{} achieves the highest Overall Consistency and PSNR and the lowest LPIPS among PTQ methods, with the best SSIM in three of four settings.
The gains are most consistent in dense-reference fidelity and Overall Consistency, while individual appearance attributes remain mixed. On Wan~2.1-1.3B at W4A4, for example, OrbitQuant retains higher image and aesthetic quality scores, whereas \method{} better preserves the dense reference across all three paired metrics. This contrast highlights that preserving the dense model's output and maximizing individual appearance scores are related but distinct goals. On Self Forcing, the fidelity gains accompany the highest Motion Smoothness at both precisions, extending the benefit to autoregressive generation.

\begin{table*}[t]
\centering
\caption{\textbf{Quantitative evaluations of scale and architecture transfer on VBench (T2V).}}
\label{tab:transfer}
\setlength{\tabcolsep}{2.2pt}
\renewcommand{\arraystretch}{1.06}
\resizebox{0.92\textwidth}{!}{%
\begin{tabular}{llc@{\hspace{5pt}}cccccccc@{\hspace{7pt}}ccc}
\toprule
\multirow[c]{2}{*}{Model} & \multirow[c]{2}{*}{Method} & \multirow[c]{2}{*}{W/A} & \multicolumn{8}{c}{VBench attributes (\%)} & \multicolumn{3}{c}{Dense-reference fidelity} \\
\cmidrule(lr){4-11}\cmidrule(lr){12-14}
 & & & IQ $\uparrow$ & AQ $\uparrow$ & MS $\uparrow$ & DD $\uparrow$ & BC $\uparrow$ & SC $\uparrow$ & Scene $\uparrow$ & OC $\uparrow$ & PSNR $\uparrow$ & SSIM $\uparrow$ & LPIPS $\downarrow$ \\
\midrule

 \multirow{7}{*}{\shortstack[c]{Wan~2.1-T2V\\(\#Param:14B)}}
 & Dense BF16 & 16/16 & 67.14 & 67.12 & 98.48 & 66.67 & 96.69 & 94.36 & 39.46 & 26.46 & ref. & ref. & ref. \\
\cmidrule(lr){2-14}
 & ViDiT-Q~\citep{zhao2024viditq} & 4/6 & \best{67.25} & 66.23 & 98.58 & 56.94 & 96.30 & 94.10 & \second{41.72} & \second{26.31} & 15.89 & 0.545 & 0.378 \\
 & OrbitQuant~\citep{lee2026orbitquant} & 4/6 & 67.18 & \second{67.14} & \second{98.60} & \best{62.50} & \second{96.44} & \second{95.01} & 37.86 & 26.22 & \best{16.93} & \best{0.585} & \best{0.325} \\
 & \cellcolor{rowgray}\method\ (ours) & \cellcolor{rowgray}4/6 & \cellcolor{rowgray}\second{67.22} & \cellcolor{rowgray}\best{67.34} & \cellcolor{rowgray}\best{98.73} & \cellcolor{rowgray}\second{61.11} & \cellcolor{rowgray}\best{96.56} & \cellcolor{rowgray}\best{95.21} & \cellcolor{rowgray}\best{42.81} & \cellcolor{rowgray}\best{26.52} & \cellcolor{rowgray}\second{16.90} & \cellcolor{rowgray}\second{0.579} & \cellcolor{rowgray}\second{0.331} \\
\cmidrule(lr){2-14}
 & ViDiT-Q~\citep{zhao2024viditq} & 4/4 & 59.41 & 58.36 & 98.13 & 30.56 & 95.00 & 91.59 & 25.29 & 23.60 & 13.27 & 0.415 & 0.567 \\
 & OrbitQuant~\citep{lee2026orbitquant} & 4/4 & \second{66.55} & \second{67.31} & \second{98.53} & \best{55.56} & \best{95.71} & \second{94.44} & \second{43.60} & \second{26.17} & \second{16.25} & \second{0.554} & \second{0.360} \\
 & \cellcolor{rowgray}\method\ (ours) & \cellcolor{rowgray}4/4 & \cellcolor{rowgray}\best{66.93} & \cellcolor{rowgray}\best{67.50} & \cellcolor{rowgray}\best{98.59} & \cellcolor{rowgray}\best{55.56} & \cellcolor{rowgray}\second{95.46} & \cellcolor{rowgray}\best{94.61} & \cellcolor{rowgray}\best{43.68} & \cellcolor{rowgray}\best{26.28} & \cellcolor{rowgray}\best{16.70} & \cellcolor{rowgray}\best{0.565} & \cellcolor{rowgray}\best{0.349} \\
 
\midrule

\multirow{7}{*}{\shortstack[c]{Wan~2.2-T2V\\(\#Param:27B)}}
 & Dense BF16 & 16/16 & 69.58 & 68.17 & 97.82 & 77.78 & 95.10 & 92.48 & 48.40 & 26.70 & ref. & ref. & ref. \\
\cmidrule(lr){2-14}
 & ViDiT-Q~\citep{zhao2024viditq} & 4/6 & \best{69.95} & \best{68.06} & \second{97.96} & 76.39 & 94.79 & \second{92.68} & \best{48.47} & 26.55 & 14.90 & 0.464 & 0.412 \\
 & OrbitQuant~\citep{lee2026orbitquant} & 4/6 & 69.23 & 67.52 & 97.95 & \second{77.78} & \best{95.03} & 92.65 & \second{47.75} & \best{26.65} & \second{15.78} & \best{0.509} & \best{0.358} \\
 & \cellcolor{rowgray}\method\ (ours) & \cellcolor{rowgray}4/6 & \cellcolor{rowgray}\second{69.86} & \cellcolor{rowgray}\second{67.93} & \cellcolor{rowgray}\best{97.98} & \cellcolor{rowgray}\best{79.17} & \cellcolor{rowgray}\second{94.80} & \cellcolor{rowgray}\best{92.81} & \cellcolor{rowgray}46.44 & \cellcolor{rowgray}\best{26.65} & \cellcolor{rowgray}\best{15.93} & \cellcolor{rowgray}\best{0.509} & \cellcolor{rowgray}\second{0.361} \\
\cmidrule(lr){2-14}
 & ViDiT-Q~\citep{zhao2024viditq} & 4/4 & \second{69.22} & 67.24 & 97.62 & 72.22 & 94.07 & 91.46 & \second{46.00} & 26.34 & 14.63 & 0.436 & 0.449 \\
 & OrbitQuant~\citep{lee2026orbitquant} & 4/4 & 68.88 & \second{67.77} & \second{97.88} & \second{75.00} & \second{94.50} & \second{92.17} & 43.82 & \best{26.72} & \second{15.15} & \second{0.476} & \second{0.396} \\
 & \cellcolor{rowgray}\method\ (ours) & \cellcolor{rowgray}4/4 & \cellcolor{rowgray}\best{69.24} & \cellcolor{rowgray}\best{67.94} & \cellcolor{rowgray}\best{97.95} & \cellcolor{rowgray}\best{76.39} & \cellcolor{rowgray}\best{94.82} & \cellcolor{rowgray}\best{92.61} & \cellcolor{rowgray}\best{46.58} & \cellcolor{rowgray}\second{26.44} & \cellcolor{rowgray}\best{15.47} & \cellcolor{rowgray}\best{0.487} & \cellcolor{rowgray}\best{0.389} \\
 
\midrule

\multirow{7}{*}{\shortstack[c]{MiniMax-H3\\(\#Param:33B)}}
 & Dense BF16 & 16/16 & 68.57 & 65.38 & 99.01 & 75.00 & 96.08 & 92.55 & 47.60 & 26.64 & ref. & ref. & ref. \\
\cmidrule(lr){2-14}
 & ViDiT-Q~\citep{zhao2024viditq} & 4/6 & 67.25 & 65.54 & 98.54 & 61.11 & \best{96.05} & \best{93.94} & 41.06 & 25.67 & 12.72 & 0.376 & 0.569 \\
 & OrbitQuant~\citep{lee2026orbitquant} & 4/6 & \second{68.35} & \second{66.20} & \second{98.97} & \best{76.39} & \second{95.54} & \second{93.27} & \second{45.64} & \best{26.54} & \second{14.04} & \second{0.420} & \second{0.499} \\
 & \cellcolor{rowgray}\method\ (ours) & \cellcolor{rowgray}4/6 & \cellcolor{rowgray}\best{68.51} & \cellcolor{rowgray}\best{66.67} & \cellcolor{rowgray}\best{98.98} & \cellcolor{rowgray}\second{75.00} & \cellcolor{rowgray}95.47 & \cellcolor{rowgray}92.50 & \cellcolor{rowgray}\best{49.56} & \cellcolor{rowgray}\second{26.26} & \cellcolor{rowgray}\best{14.26} & \cellcolor{rowgray}\best{0.430} & \cellcolor{rowgray}\best{0.496} \\
\cmidrule(lr){2-14}
 & ViDiT-Q~\citep{zhao2024viditq} & 4/4 & 59.93 & 53.94 & 97.82 & 43.06 & \second{95.08} & 91.10 & 20.35 & 22.99 & 12.57 & 0.349 & 0.646 \\
 & OrbitQuant~\citep{lee2026orbitquant} & 4/4 & \second{66.48} & \second{65.35} & \second{98.72} & \second{70.83} & \best{95.16} & \best{92.84} & \second{42.30} & \second{26.14} & \second{13.86} & \second{0.409} & \second{0.518} \\
 & \cellcolor{rowgray}\method\ (ours) & \cellcolor{rowgray}4/4 & \cellcolor{rowgray}\best{66.83} & \cellcolor{rowgray}\best{65.43} & \cellcolor{rowgray}\best{98.73} & \cellcolor{rowgray}\second{70.83} & \cellcolor{rowgray}94.97 & \cellcolor{rowgray}\second{92.23} & \cellcolor{rowgray}\best{46.88} & \cellcolor{rowgray}\best{26.24} & \cellcolor{rowgray}\best{14.17} & \cellcolor{rowgray}\best{0.424} & \cellcolor{rowgray}\best{0.514} \\
\bottomrule
\end{tabular}}
\end{table*}

\begin{figure}[t]
    \centering
    \includegraphics[width=\linewidth]{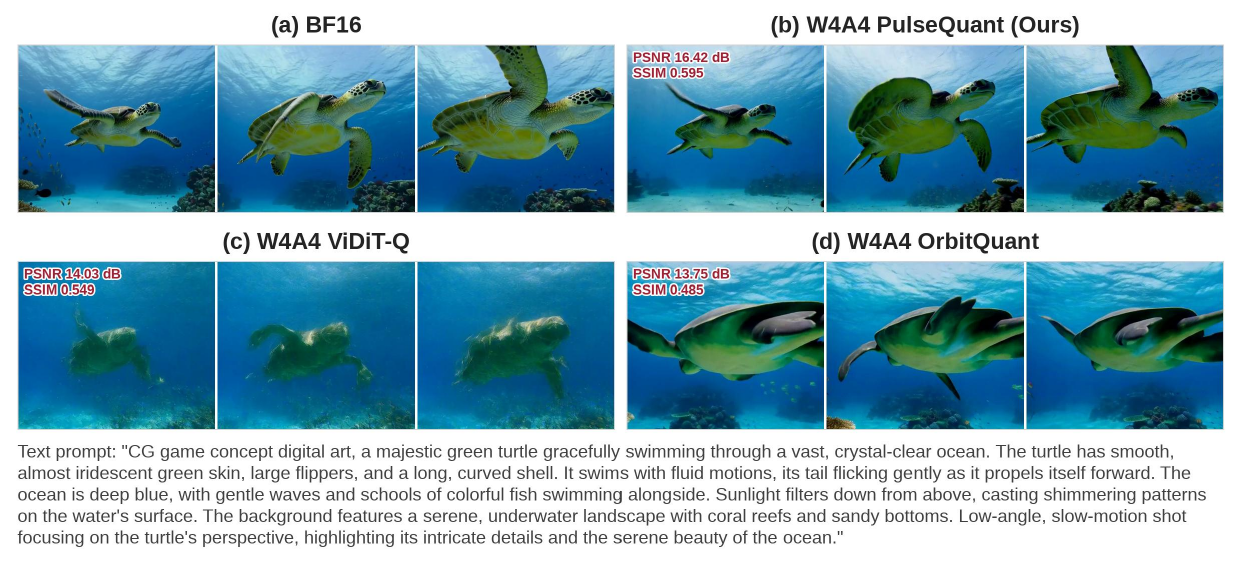}
    \vspace{-0.37in}
    \caption{MiniMax-H3 W4A4 comparison at frames 5, 60, and 100 under the same condition. }
    \label{fig:h3-qualitative}
\end{figure}

\noindent\textbf{Transfer across scale and architecture.}
Table~\ref{tab:transfer} evaluates transfer across models with 14B--33B parameters. At W4A4, \method{} achieves the best PSNR, SSIM, and LPIPS among PTQ methods on all three models, spanning bidirectional Wan models and autoregressive MiniMax-H3.
At W4A6, results are more mixed: \method{} leads six VBench attributes on Wan~2.1-14B, while OrbitQuant retains the best paired fidelity; on MiniMax-H3, \method{} leads all paired metrics, but OrbitQuant achieves higher Overall Consistency. The clearest cross-model advantage is therefore dense-reference fidelity at W4A4, with more metric-dependent trade-offs at W4A6.
Figure~\ref{fig:h3-qualitative} presents a qualitative MiniMax-H3 comparison under matched generation conditions. 
ViDiT-Q preserves the coarse layout but suppresses fine texture, whereas OrbitQuant remains sharp while changing the subject scale and trajectory; \method{} more closely follows the dense composition and detail. 
Appendix~\ref{app:qualitative} provides further qualitative comparisons across models and precisions.



\subsection{Ablation Studies}
\label{sec:ablations}
We test the propagation premise, isolate the two calibration components, and analyze correction and profiling choices. Complete sweeps and secondary controls are reported in Appendix~\ref{app:additional-ablations}.

\begin{table*}[t]
\centering
\begin{minipage}[t]{0.49\textwidth}
\centering
\captionof{table}{Component ablation on Wan~2.1-T2V-1.3B W4A6 with $\lambda=0.5$ and $\rho=0.25$.}
\label{tab:ablation}
\scriptsize
\setlength{\tabcolsep}{3.6pt}
\renewcommand{\arraystretch}{1.08}
\begin{tabular}{@{}lccccc@{}}
\toprule
Configuration & IQ $\uparrow$ & AQ $\uparrow$ & OC $\uparrow$ & PSNR $\uparrow$ & LPIPS $\downarrow$ \\
\midrule
Uniform PTQ       & 56.90 & 57.29 & 23.43 & 14.00 & 0.549 \\
$+$ Rotation      & 63.20 & 62.39 & 24.70 & 14.97 & 0.443 \\
$+$ Risk          & 64.38 & \best{64.97} & 25.73 & 15.59 & 0.414 \\
$+$ Subspace corr.& \second{64.64} & 64.24 & \second{25.74} & \second{15.61} & \second{0.412} \\
\rowcolor{rowgray} \method & \best{65.83} & \second{64.93} & \best{26.18} & \best{16.08} & \best{0.385} \\
\bottomrule
\end{tabular}

\end{minipage}\hfill
\begin{minipage}[t]{0.49\textwidth}
\centering
\captionof{table}{Ablation results of correction-direction ablation on Wan~2.1-T2V-1.3B W4A4 with $\lambda=0.5$ and $\rho=0.25$.}
\vspace{0.03in}
\label{tab:direction-control}
\scriptsize
\setlength{\tabcolsep}{5.0pt}
\renewcommand{\arraystretch}{0.96}
\begin{tabular}{@{}lccccc@{}}
\toprule
Direction & IQ $\uparrow$ & AQ $\uparrow$ & DD $\uparrow$ & Scene $\uparrow$ & OC $\uparrow$ \\
\midrule
None                      & \second{62.17} & \best{62.90}   & \second{51.39} & 35.17           & \second{25.50} \\
Random                    & 61.62           & \second{62.86} & 48.61           & \second{37.72} & 25.38 \\
Fixed Walsh               & 61.96           & 62.24           & \second{51.39} & 34.52           & 25.08 \\
\rowcolor{rowgray}
Calibrated (ours)         & \best{62.70}   & 62.46           & \best{55.56}   & \best{37.78}   & \best{25.64} \\
\bottomrule
\end{tabular}

\end{minipage}
\vspace{-0.15in}
\end{table*}

\paragraph{Effect of each component.}
Table~\ref{tab:ablation} separates the contributions of rotation and the two calibration modules. The risk-only and correction-only variants independently extend the same rotation baseline; the full method combines both modules. Rotation provides the initial quality recovery, and each calibration module further improves Overall Consistency and paired fidelity. Combining both modules gives the best Overall Consistency, PSNR, and LPIPS, supporting their complementary roles in trajectory prioritization and response preservation. Table~\ref{tab:ablation-full} further shows that subspace correction reduces high-pass temporal error, linking response preservation to temporal fidelity.

\begin{figure*}[t]
\centering
\includegraphics[width=\textwidth]{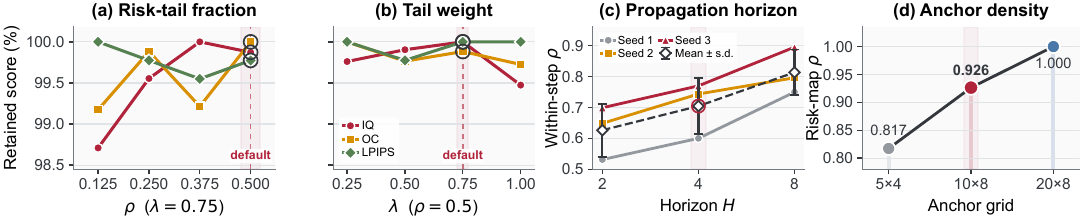}
\vspace{-0.3in}
\caption{Sensitivity and profiling choices. (a--b) Risk-objective robustness on Wan~2.1-T2V-1.3B W4A4. (c--d) Propagation-horizon stability and anchor-grid agreement on Wan~2.1-T2V-14B.}
\vspace{-0.15in}
\label{fig:sensitivity-profiling}
\end{figure*}


\begin{wrapfigure}[10]{r}{0.54\linewidth}
\vspace{-0.7\baselineskip}
\centering
\includegraphics[width=\linewidth]{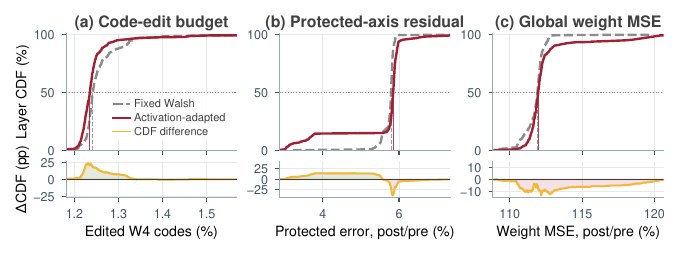}
\vspace{-0.35in}
\caption{Response-axis adaptation lowers subspace error with code edits and weight MSE in Wan-14B.}
\label{fig:layer-audit}
\end{wrapfigure}

\paragraph{Correction analysis.}
Activation-adapted directions improve correction quality at a comparable edited-code fraction. In Table~\ref{tab:direction-control}, calibrated axes lead in four of five metrics over random, fixed Walsh, and no-correction controls. Figure~\ref{fig:layer-audit} shows lower protected-subspace residuals with nearly unchanged edited-code fraction and global weight MSE. Together, these results support directing code edits toward calibration-relevant responses rather than minimizing weight error alone.

\paragraph{Effect of propagation risk.}
Figure~\ref{fig:propagation-evidence} exposes the limitation of local reconstruction: block-output RMSE is nearly uncorrelated with final latent error ($\rho=-0.08$), whereas four-step propagated error recovers the within-timestep ranking of final error ($\rho=0.70$). Perturbations of similar local magnitude can therefore have different downstream effects, supporting propagation-aware calibration.

\paragraph{Sensitivity and profiling design.}
Figure~\ref{fig:sensitivity-profiling} examines sensitivity to the risk-objective parameters and profiling granularity. 
Performance remains stable over the tested $\rho$ and $\lambda$ ranges. We use $\rho=0.5$ and $\lambda=0.75$ to balance average response error with emphasis on calls having large risk-weighted errors.
For propagation profiling, $H=4$ maintains a strong correlation with final error across seeds while requiring fewer continuation steps than $H=8$.
The $10\!\times\!8$ block--timestep grid also closely agrees with the $20\!\times\!8$ reference while halving the number of profiling calls. These settings provide the default balance between profiling fidelity and cost.

\subsection{Efficiency Discussion}
\label{sec:efficiency}
\begin{wrapfigure}{r}{0.67\linewidth}
    \centering
    \vspace{-0.15in}
    \setlength{\abovecaptionskip}{0.2cm}
    \includegraphics[width=\linewidth]{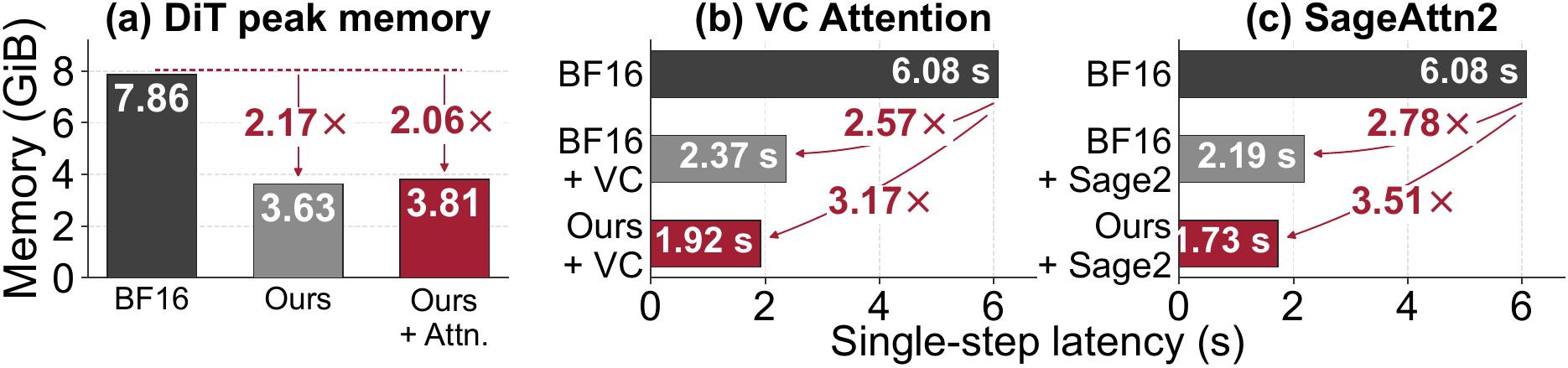}
    \vspace{-0.18in}
    \caption{RTX~5080 W4A4 efficiency: (a) DiT memory; (b,c) VC/SageAttention2 per-step latency and speedup relative to BF16.}
    \label{fig:deployment-efficiency}
    \vspace{-0.15in}
\end{wrapfigure}

\noindent\textbf{Native inference efficiency.}
Figure~\ref{fig:deployment-efficiency} reports native W4A4 inference on Wan~2.1-1.3B using one RTX~5080 under the measurement protocol above. Appendix~\ref{app:deployment-details} details the packed NVFP4 implementation and kernels.

Packed NVFP4 reduces DiT peak memory from 7.86 to 3.63~GiB; combining it with low-bit attention brings the peak to 3.81~GiB.
Low-bit linear layers also complement the latency savings from low-bit attention.
VC-Attention~\citep{li2026vcattention} reduces DiT-step latency from 6.08 to 2.37~s; combining it with \method{} further lowers latency to 1.92~s, giving a 3.17$\times$ speedup over Dense BF16.
SageAttention2~\citep{zhang2024sageattention2} reaches 2.19~s alone and 1.73~s with \method{}, for a 3.51$\times$ speedup over the same baseline.
Thus, native W4A4 linear kernels provide additional reductions in DiT-step latency on top of both low-bit attention backends.

\section{Conclusion}
We introduced \method{}, a 4-bit post-training quantization method that combines propagation-guided radius calibration with activation-informed code correction. It uses downstream sensitivity to prioritize calibration states and dominant activation directions to guide discrete weight updates, preserving the existing 4-bit representation. Controlled interventions show that short-horizon propagated error predicts final latent error more reliably than immediate block-output error. Evaluations across bidirectional and autoregressive video generators demonstrate gains in key quality metrics, with consistent dense-reference fidelity advantages over the state-of-the-art PTQ baselines. Native packed NVFP4 deployment reduces DiT memory and provides additional DiT-step acceleration when combined with low-bit attention. These results support calibration guided by both error propagation and activation response geometry. Reducing profiling cost and extending correction beyond a compact activation subspace remain important directions for future work.

\clearpage
\section*{Reproducibility Statement}
Section~\ref{sec:method} describes the calibration procedure and discrete correction algorithm, with derivations in Appendix~\ref{app:derivations}. Section~\ref{sec:experiments} specifies the evaluated models, baselines, metrics, and comparison protocol; Appendix~\ref{app:protocol} details the quantization operators, profiling configuration, radius search, and default hyperparameters. Quality comparisons use matched prompts, seeds, and scheduler settings with quantization--dequantization on H200, while native deployment is evaluated separately on RTX~5080. Appendix~\ref{app:deployment-details} documents kernel coverage, software versions, and measurement boundaries for the systems results.
\section*{Ethics Statement}
\method{} lowers the memory and computational requirements of video generation, enabling broader access to existing models on resource-constrained hardware. These efficiency gains can support research and creative applications with more modest computing resources. Our experiments evaluate the quality and efficiency of quantized models and involve no human-subject studies. Responsible use should follow the licenses and safeguards of the underlying models, respect privacy and consent, and clearly disclose synthetic content where appropriate to mitigate misleading or harmful uses.
\section*{AI Use Statement}
AI tools assisted with research ideation and execution, including discussing methodological alternatives, supporting code implementation and debugging, and inspecting experimental records. 
They also helped organize, revise, and polish the manuscript for clarity and readability. 
All AI-assisted text and analysis were reviewed by the authors, and every reported number, claim, and citation was verified against our own experimental logs and the cited sources. We take responsibility for the final content of this work.

\bibliography{walsh2_references}
\bibliographystyle{iclr2027_conference}

\clearpage
\appendix
\begin{center}
  {\Large\bfseries Appendix\par}
\end{center}
\vspace{0.65em}

\begingroup
\hypersetup{hidelinks}
\setlength{\fboxsep}{8pt}
\setlength{\fboxrule}{0.4pt}
\noindent\fcolorbox{black}{white}{%
\begin{minipage}{\dimexpr\linewidth-2\fboxsep-2\fboxrule\relax}
\raggedright
{\normalsize\scshape Contents\par}
\vspace{0.45em}
\small
\setlength{\parskip}{0.12em}
\newcommand{\appcontentsentry}[5]{%
  \noindent\hspace*{#1}%
  {\color{appendixred}%
  \hyperref[#2]{%
    \makebox[#3][l]{#4}#5%
  }}%
  \nobreak\leaders\hbox to 0.55em{\hss.\hss}\hfill\nobreak
  \pageref{#2}\par
}
\newcommand{\appsectionentry}[3]{%
  \noindent{\color{appendixred}\hyperref[#1]{%
    \makebox[2.2em][l]{\textbf{#2}}\textbf{\MakeUppercase{#3}}}}%
  \hfill\textbf{\pageref{#1}}\par
}
\newcommand{\appsubsectionentry}[3]{%
  {\footnotesize\appcontentsentry{1.2em}{#1}{3.2em}{#2}{\MakeUppercase{#3}}}%
}

\appsectionentry{app:derivations}{A}{Method derivations and discrete correction}
\appsubsectionentry{app:propagation-first-order}{A.1}{First-order view of propagation risk}
\appsubsectionentry{app:response-error}{A.2}{Response-error decomposition}
\appsubsectionentry{app:discrete-search}{A.3}{Adjacent-code prefix search}
\appsectionentry{app:protocol}{B}{Experimental protocol and implementation details}
\appsubsectionentry{app:quantization-implementation}{B.1}{Quantization and calibration implementation}
\appsubsectionentry{app:default-hyperparameters}{B.2}{Default hyperparameter settings}
\appsubsectionentry{app:notation}{B.3}{Notation}
\appsectionentry{app:additional-ablations}{C}{Additional ablation studies}
\appsubsectionentry{app:component-full}{C.1}{Complete component results}
\appsubsectionentry{app:calibration-prompts}{C.2}{Calibration-set sensitivity}
\appsubsectionentry{app:risk-hyperparameters}{C.3}{Propagation-risk objective}
\appsectionentry{app:deployment-details}{D}{Native W4A4 deployment and profiling}
\appsubsectionentry{app:packed-linear}{D.1}{Packed linear path}
\appsubsectionentry{app:fusion-workspace}{D.2}{Fusion and workspace reuse}
\appsubsectionentry{app:attention-measurement}{D.3}{Attention backend and measurement boundary}
\appsectionentry{app:qualitative}{E}{Additional qualitative results}
\appsubsectionentry{app:qual-wan13-w4a4}{E.1}{Wan 2.1-1.3B W4A4 comparisons}
\appsubsectionentry{app:qual-transfer}{E.2}{W4A6 and cross-architecture comparisons}
\end{minipage}%
}
\endgroup
\vspace{0.75em}

\section{Method derivations and discrete correction}
\label{app:derivations}
This section provides the derivations omitted from the main text. We first relate the measured propagation gain to first-order trajectory dynamics, then decompose the response error controlled by subspace correction, and finally specify the legal-code search used by \method{}.

\subsection{First-order view of propagation risk}
\label{app:propagation-first-order}
Write one dense denoising update as $\mathbf z_{t+1}=F_t(\mathbf z_t)$, and let $\mathbf e_t$ be a small perturbation of the trajectory state. Linearizing each update gives
\begin{equation}
 \mathbf e_{t+1}\approx\mathbf J_t\mathbf e_t+\mathbf u_t,
 \qquad
 \mathbf e_T\approx\sum_{t=0}^{T-1}\boldsymbol\Phi_{T,t+1}\mathbf u_t,
 \quad
 \boldsymbol\Phi_{T,t+1}=\mathbf J_{T-1}\cdots\mathbf J_{t+1},
 \label{eq:app-propagation}
\end{equation}
where $\mathbf J_t$ is the Jacobian of $F_t$ and $\mathbf u_t$ is the local error introduced at update $t$. For an isolated error pulse, the ratio of its propagated error to its immediate error estimates the finite-horizon amplification induced by $\boldsymbol\Phi_{t+1+H,t+1}$. Equation~\ref{eq:gain} measures this amplification directly from latent trajectories and therefore does not require explicit Jacobian construction.

\subsection{Response-error decomposition}
\label{app:response-error}
We first consider the full layer-input space, where the two response axes are orthonormal. Write $\x'=\beta_0\mathbf a_0+\beta_1\mathbf a_1+\x_\perp$, with $\beta_u=\langle\x',\mathbf a_u\rangle$ and $\x_\perp$ orthogonal to both axes. For the full-row residual $\mathbf r=\mathbf w'_i-\mathbf q_i$, the triangle and Cauchy--Schwarz inequalities give
\begin{equation}
 |\langle\mathbf r,\x'\rangle|
 \leq
 \sum_{u=0}^{1}|\beta_u|\,|\langle\mathbf r,\mathbf a_u\rangle|
 +\|\mathbf r\|_2\|\x_\perp\|_2.
 \label{eq:app-response-bound}
\end{equation}
The bound separates response error along the selected axes from the contribution outside their span. Estimating these axes from transformed calibration activations focuses correction on directions with high observed activation energy, while controlling the residual norm limits its interaction with the remaining component.

Equation~\ref{eq:joint-correction} implements this principle group-wise. Each full-row response residual decomposes as $\langle\mathbf r,\mathbf a_u\rangle=\sum_{\mathcal G}\langle\mathbf r_{\mathcal G},\mathbf a_{u,\mathcal G}\rangle$, where subscripts $\mathcal G$ denote channel slices. These slices generally have different norms and need not remain mutually orthogonal. The group objective therefore uses normalized directional penalties, together with full-group distortion, as a local surrogate for response preservation rather than an exact orthogonal decomposition of the full-row error.

\subsection{Adjacent-code prefix search}
\label{app:discrete-search}
For a coordinate with current code $c_j$, the two legal adjacent moves and their induced weight changes are
\begin{equation}
 c_j^- = \max(c_j-1,0),\quad
 c_j^+ = \min(c_j+1,15),\quad
 \delta_j^s=\bar r_i(C_{c_j^s}-C_{c_j}),\quad s\in\{-,+\}.
 \label{eq:app-neighbor}
\end{equation}
All single-coordinate proposals in a correction pass are evaluated from the same current reconstruction $\mathbf q$. For each coordinate, we retain the sign $s_j$ that produces the smaller single-move value of $\mathcal L_{i,\mathcal G}$. We sort the retained proposals by increasing objective change to obtain an order $\pi$. For every cumulative prefix, we recompute the full joint objective, including the cross terms induced by simultaneous coordinate changes in the protected responses, and select
\begin{equation}
 K^*=\arg\min_{K\in\{0,\ldots,G\}}
 \mathcal L_{i,\mathcal G}\!\left(
 \mathbf q+\sum_{k=1}^{K}\delta_{\pi(k)}^{s_{\pi(k)}}
 \mathbf e_{\pi(k)}\right).
 \label{eq:app-prefix}
\end{equation}
The empty prefix $K=0$ is evaluated first, and ties are resolved in favor of the earliest prefix. Thus, the original reconstruction is retained whenever no non-empty prefix improves the objective. Including $K=0$ ensures $\mathcal L_{i,\mathcal G}(\mathbf q_{\mathrm{new}})\leq\mathcal L_{i,\mathcal G}(\mathbf q)$. Every accepted coordinate uses a legal codebook index at the calibrated row radius, preserving the deployment format.

\section{Experimental protocol and implementation details}
\label{app:protocol}
This section specifies the quantization operators and calibration settings used in the experiments. Within each setting, methods share the model revision, quantized-layer coverage, evaluation prompts, seeds, scheduler, and evaluator. Calibration statistics are re-estimated for each checkpoint and precision.

\subsection{Quantization and calibration implementation}
\label{app:quantization-implementation}
\paragraph{Activation quantizer.}
For unclipped symmetric per-token quantization with activation bit width $b_a$, let $q_{\max}=2^{b_a-1}-1$ and $\alpha=\max(\|\x'\|_\infty/q_{\max},\epsilon_{\mathrm{scale}})$. The dequantized output is
\begin{equation}
Q_A(\x')=\alpha\,\operatorname{clip}\big(\operatorname{round}(\x'/\alpha),-q_{\max},q_{\max}\big).
\end{equation}
Reduction and rounding are performed in FP32 before restoring the input dtype. The scale floor, rounding rule, and backend are held fixed across methods within each comparison.

\paragraph{Profiling pulse configuration.}
\label{app:profiling-pulse}
The CFG-based block interventions used to construct the propagation-risk profiles apply a profiling-only uniform W4A4 quantizer to ten principal linear projections: the Q, K, V, and output projections of self-attention and cross-attention, together with the FFN input and output projections. Weights use channel balancing, block-Hadamard coordinates, and groupwise calibrated uniform 4-bit quantization. Activations use dynamic per-token symmetric 4-bit quantization with a frozen clipping schedule. The pulse is active in both CFG forward calls at the selected timestep; all other blocks and the subsequent $H$ denoising updates execute in dense BF16.

\paragraph{Radius candidate grid.}
\label{app:radius-grid}
For each transformed weight row, we initialize the radius as $r_i^{(0)}=\|\mathbf w'_i\|_2$ and construct a symmetric multiplicative grid:
\begin{equation}
\mathcal R_i=\left\{r_i^{(0)}\left(1-\eta+\frac{2\eta k}{K_r-1}\right)\right\}_{k=0}^{K_r-1}.
\label{eq:app-radius-grid}
\end{equation}
The main experiments on Wan~2.1-1.3B, Wan~2.1-14B, MiniMax-H3, and Self Forcing use $K_r=5$ and $\eta=0.08$, giving $\mathcal R_i=r_i^{(0)}\{0.92,0.96,1.00,1.04,1.08\}$. The grid searches relative changes around each row norm rather than a shared absolute offset.

\paragraph{Adaptation to autoregressive generation.}
\label{app:autoregressive-adaptation}
For Self Forcing, earlier generated chunks provide cached context for later chunks. We therefore adapt the calibration weights to this causal execution order using a local response-energy proxy and a remaining-context factor, in place of the isolated-pulse profile in Eq.~\ref{eq:gain}. The calibration trajectory contains seven temporal chunks, each with four denoising calls and one clean-context KV-cache update, giving 35 calls per prompt for the repeatedly executed projections. We collect dense activations from three complementary prompts, retaining four tokens per call.

For layer $\ell$, prompt $p$, and call $s$, let $\mathbf U_{\ell,p,s}$ denote the sampled layer inputs before balancing and rotation, and let $\mathbf V_{\ell,p,s}=\mathbf U_{\ell,p,s}\W_\ell^\top$ be their bias-free linear responses. We compute
\begin{equation}
 \begin{aligned}
 a_{\ell,p,s}&=\sqrt{\frac{\max\{\operatorname{Mean}(\mathbf V_{\ell,p,s}^{\odot 2}),\epsilon\}}
 {\max\{\operatorname{Mean}(\mathbf U_{\ell,p,s}^{\odot 2}),\epsilon\}}},\\
 \omega^{\mathrm{AR}}_{\ell,s}&=\operatorname{MeanOne}_{s}\!\left[\bar a_{\ell,s}(7-\lfloor s/5\rfloor)^{1/2}\right],
 \end{aligned}
 \label{eq:ar-calibration-weights}
\end{equation}
where the energy means run over tokens and channels, $s\in\{0,\ldots,34\}$, and $\epsilon=10^{-12}$. We normalize the gains by their geometric mean over calibration calls and aggregate corresponding calls across prompts by a geometric mean to obtain $\bar a_{\ell,s}$. Starting from a uniform prior, the final weights have unit arithmetic mean over calls within each layer. The causal factor counts the current and remaining chunks, assigning greater weight to earlier context. Thus, these weights combine measured local response energy with a causal-position heuristic rather than a finite-horizon perturbation measurement.

Cross-attention key and value projections are evaluated once per prompt and cached; they use a single state with unit normalized weight and no remaining-chunk factor. We substitute $\omega^{\mathrm{AR}}_{\ell,s}$ for $\omega_{\ell,t(s)}$ in Eq.~\ref{eq:radius-objective}, with $\lambda=0.75$ and $\rho=0.25$. Row-radius search and response-subspace correction otherwise follow the same procedure. All weights are calibrated before evaluation, with no online risk estimation or code correction.

\subsection{Default hyperparameter settings}
\label{app:default-hyperparameters}
Table~\ref{tab:default-configuration} summarizes the shared calibration configuration. We use three complementary prompts covering object appearance, object motion, and camera motion; Appendix~\ref{app:calibration-prompts} reports the corresponding prompt-composition study. These defaults are fixed across the main experiments unless a model-specific exception is stated in the evaluation protocol.

\begin{table}[t]
\centering
\caption{Default \method{} calibration configuration. Model-specific exceptions are reported with the evaluation protocol.}
\label{tab:default-configuration}
\small
\setlength{\tabcolsep}{4pt}
\renewcommand{\arraystretch}{1.06}
\begin{tabularx}{\columnwidth}{@{}l>{\raggedright\arraybackslash}Xl@{}}
\toprule
Stage & Setting & Default \\
\midrule
Calibration data & Number of prompts & 3 \\
Calibration data & Prompt composition & Complementary coverage \\
Propagation profile & Horizon $H$ & 4 steps \\
Propagation profile & Anchor grid & $10$ blocks $\times$ $8$ timesteps \\
Risk objective & Tail weight $\lambda$ & 0.75 \\
Risk objective & Tail fraction $\rho$ & 0.5 \\
Radius search & Number of candidates $K_r$ & 5 \\
Radius search & Relative half-span $\eta$ & 0.08 \\
Response axes & Token budget $M_{\max}$ & 512 \\
Response axes & Power iterations $K_{\mathrm p}$ & 2 \\
Response axes & Number of protected axes & 2 \\
Code correction & Group width $G$ & 128 \\
Code correction & Distortion weight $\tau$ & 0.25 \\
\bottomrule
\end{tabularx}
\end{table}

\subsection{Notation}
\label{app:notation}
Table~\ref{tab:notation} lists the symbols used in the method and appendices.

\par\medskip\noindent
\begin{minipage}{\linewidth}
\centering\small
\captionof{table}{Notation used in the method and implementation details.}
\label{tab:notation}
\begin{tabularx}{\linewidth}{@{}l>{\raggedright\arraybackslash}X@{}}
\toprule Symbol & Meaning\\\midrule
$\W,\x,\X;m,d,N$ & Weight matrix, token, batch; output/input widths, tokens\\
$\Rot,\D$ & Orthogonal RPBH transform; positive channel scaling\\
$\mathcal C,C_k,c_{ij}$ & Shared 16-value codebook, entry, and coordinate index\\
$Q_A,b_a$ & Activation quantizer and activation bit width\\
$r_i^{(0)},r_i,\bar r_i$ & Initial row norm (also denoted $r_i$) and calibrated radius\\
$\mathcal R_i,K_r,\eta$ & Radius candidates, grid size, and relative half-span\\
$\ell,i,j;s,t(s)$ & Layer, row, coordinate; calibration call and its denoising step\\
$t,T,H$ & Denoising step, total updates, and profiling horizon\\
$\X'_{\ell,s},N_s$ & Transformed calibration activations and token count at call $s$\\
$g_{b,t},\bar g_{\ell,t},\omega_{\ell,t}$ & Anchor gain, interpolated gain, and normalized calibration weight\\
$\gamma,[a,b]$ & Gain exponent and clipping interval\\
$e_{i,s}(r),u_{i,s}(r)$ & Row-output error and its risk-weighted value\\
$\lambda,\rho$ & Tail-term weight and retained fraction of calibration calls\\
$\mathbf p_u,\mathbf a_{\ell,u}$ & Fixed native-input probe and activation-adapted response axis\\
$\bar\X'_\ell,M_\ell,M_{\max}$ & Retained activations, retained token count, and token budget\\
$K_{\mathrm p},\mathbf C_\ell$ & Power-iteration count and uncentered second moment\\
$\mathcal G,G;\mathbf w,\mathbf q$ & Channel group and width; dense and quantized group slices\\
$\delta_j^s,\tau$ & Weight change for move $s\in\{-,+\}$ and full-group distortion weight\\\bottomrule
\end{tabularx}
\end{minipage}
\par\medskip

\section{Additional ablation studies}
\label{app:additional-ablations}
This section supplements the compact ablations in the main paper with complete metrics and controlled studies of the calibration data and propagation-risk objective.

\subsection{Complete component results}
\label{app:component-full}
Table~\ref{tab:ablation-full} reports the complete metrics corresponding to Table~\ref{tab:ablation}. All variants quantize the same 300 linear layers under an identical W4A6 protocol. Their differences therefore isolate the effects of rotation, propagation weighting, and response-subspace correction.

\par\medskip\noindent
\begin{minipage}{\linewidth}
\centering
\captionof{table}{Complete component ablation on Wan~2.1-T2V-1.3B W4A6 under matched 300-layer coverage.}
\label{tab:ablation-full}
\setlength{\tabcolsep}{2.8pt}
\renewcommand{\arraystretch}{1.06}
\resizebox{1.0\textwidth}{!}{%
\begin{tabular}{lccc@{\hspace{7pt}}ccccc@{\hspace{7pt}}cccc}
\toprule
& \multicolumn{3}{c}{Components} & \multicolumn{5}{c}{VBench attributes (\%)} & \multicolumn{4}{c}{Dense-reference fidelity} \\
\cmidrule(lr){2-4}\cmidrule(lr){5-9}\cmidrule(lr){10-13}
Configuration & Rotation & Risk & Subsp. corr. & IQ $\uparrow$ & AQ $\uparrow$ & MS $\uparrow$ & BC $\uparrow$ & OC $\uparrow$ & PSNR $\uparrow$ & SSIM $\uparrow$ & LPIPS $\downarrow$ & T-HP $\downarrow$ \\
\midrule
Uniform PTQ & & & & 56.90 & 57.29 & 97.38 & 95.01 & 23.43 & 14.00 & 0.421 & 0.549 & 0.161 \\
Rotation only & $\checkmark$ & & & 63.20 & 62.39 & 98.09 & 95.01 & 24.70 & 14.97 & 0.512 & 0.443 & 0.155 \\
Risk only & $\checkmark$ & $\checkmark$ & & 64.38 & \best{64.97} & 98.18 & \second{95.58} & 25.73 & 15.59 & 0.529 & 0.414 & 0.150 \\
Subspace corr. only & $\checkmark$ & & $\checkmark$ & \second{64.64} & 64.24 & \second{98.26} & \best{95.69} & \second{25.74} & \second{15.61} & \second{0.536} & \second{0.412} & \best{0.142} \\
\rowcolor{rowgray} \method & $\checkmark$ & $\checkmark$ & $\checkmark$ & \best{65.83} & \second{64.93} & \best{98.47} & 95.30 & \best{26.18} & \best{16.08} & \best{0.546} & \best{0.385} & \second{0.144} \\
\bottomrule
\end{tabular}}
\end{minipage}
\par\medskip

Rotation provides the initial quality recovery. Relative to this baseline, risk weighting and response-subspace correction each improve Overall Consistency and all three paired fidelity metrics. The full method achieves the best PSNR, SSIM, and LPIPS, whereas correction alone yields the lowest temporal high-pass error. Thus, combining trajectory prioritization with response correction improves aggregate fidelity, while the temporal diagnostic retains a small trade-off.

\paragraph{Fixed-axis control.}
The fixed Walsh variant in Table~\ref{tab:direction-control} transports native input probes $\mathbf p_u$ into quantizer coordinates:
\begin{equation}
 \mathbf a^{\mathrm{fixed}}_{\ell,u}=\Rot\D\mathbf p_u.
 \label{eq:fixed-axes}
\end{equation}
The corresponding row-response residual is $(\mathbf w'_i-\mathbf q_i)^\top\mathbf a^{\mathrm{fixed}}_{\ell,u}$. Activation-adapted axes are instead estimated directly from transformed calibration activations $\X'$. This comparison contrasts fixed input probes with data-adapted response directions within the same correction framework.

\begin{figure*}[t]
    \centering
    \includegraphics[width=\textwidth]{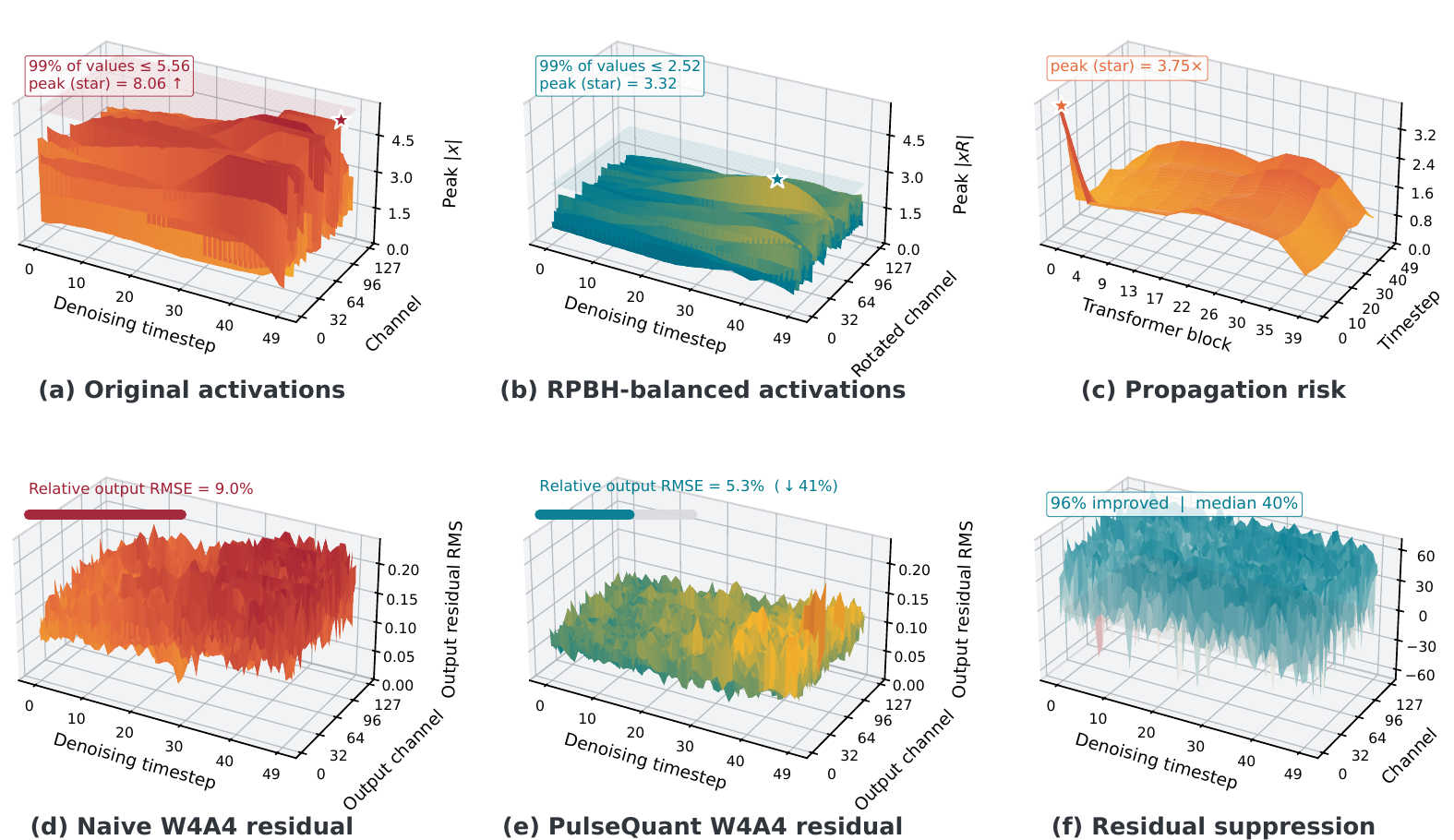}
    \caption{Layer-level activation and error diagnostics on Wan~2.1-14B W4A4 over one matched 50-step calibration trajectory. (a,b) Per-channel peak activation magnitudes before and after scaling and RPBH for the most outlier-heavy 128-channel group; planes and stars mark the 99th percentile and maximum, respectively. (c) Four-step propagation gains measured on $10$ blocks $\times$ $8$ timestep anchors; intermediate timesteps are interpolated only for visualization. (d,e) RMS output-response residuals relative to BF16 for naive W4A4 and \method{}, evaluated on 128 sampled output channels with a shared scale; bars report relative RMSE. (f) Pointwise residual reduction, where positive values favor \method{}. At this audited layer, relative RMSE decreases from 9.0\% to 5.3\%, and 96\% of channel--timestep locations improve.}
    \label{fig:trajectory-diagnostics}
\end{figure*}

\paragraph{Trajectory-wide diagnostics.}
Figure~\ref{fig:trajectory-diagnostics} examines the self-attention output projection \texttt{blocks.0.attn1.to\_out.0} on Wan~2.1-14B W4A4. The panels show activation magnitudes before and after balancing, block--timestep variation in propagation gain, and output residuals under naive quantization and \method{}. At this layer, \method{} lowers relative output RMSE by 41\%, with improvements at 96\% of the sampled channel--timestep locations. These diagnostics illustrate the activation and trajectory structure motivating calibration and complement the component comparisons in Tables~\ref{tab:ablation} and~\ref{tab:ablation-full}.

\par\medskip\noindent
\begin{minipage}{\linewidth}
\centering
\captionof{table}{Propagation-risk sensitivity on Wan~2.1-T2V-1.3B W4A4. Gray rows mark the default $(\lambda,\rho)=(0.75,0.5)$.}
\label{tab:wan13-hparam-sensitivity}
\setlength{\tabcolsep}{2.0pt}
\renewcommand{\arraystretch}{1.06}
\resizebox{\textwidth}{!}{%
\begin{tabular}{cc@{\hspace{5pt}}ccccccccc@{\hspace{7pt}}ccc@{\hspace{7pt}}c}
\toprule
\multirow[c]{2}{*}{$\lambda$} & \multirow[c]{2}{*}{$\rho$} & \multicolumn{9}{c}{VBench attributes (\%)} & \multicolumn{3}{c}{Dense-reference fidelity} & \multirow[c]{2}{*}{N-Jerk $\downarrow$} \\
\cmidrule(lr){3-11}\cmidrule(lr){12-14}
 & & IQ $\uparrow$ & AQ $\uparrow$ & MS $\uparrow$ & DD $\uparrow$ & BC $\uparrow$ & SC $\uparrow$ & Scene $\uparrow$ & OC $\uparrow$ & TF $\uparrow$ & PSNR $\uparrow$ & SSIM $\uparrow$ & LPIPS $\downarrow$ & \\
\midrule


\multicolumn{15}{l}{\textit{$\rho$ sweep with $\lambda=0.75$}} \\
0.75 & 0.125 & 61.97 & 62.03 & 97.78 & 51.39 & 94.62 & \best{92.02} & \best{38.15} & 25.23 & \best{98.60} & \best{15.45} & \best{0.506} & \best{0.442} & \second{1.429} \\
0.75 & 0.250 & 62.50 & \second{62.40} & \second{97.80} & 52.78 & 94.73 & 91.89 & 35.61 & \second{25.41} & \second{98.59} & \second{15.43} & \second{0.503} & \second{0.443} & \second{1.429} \\
0.75 & 0.375 & \best{62.78} & 62.37 & 97.79 & \best{56.94} & \best{94.91} & 91.76 & 35.03 & 25.24 & 98.57 & 15.38 & \second{0.503} & 0.444 & \second{1.429} \\
\rowcolor{rowgray} 0.75 & 0.500 & \second{62.70} & \best{62.46} & \best{97.83} & \second{55.56} & \second{94.84} & \second{91.97} & \second{36.70} & \best{25.44} & 98.57 & 15.42 & \second{0.503} & \second{0.443} & \best{1.427} \\
\midrule

\multicolumn{15}{l}{\textit{$\lambda$ sweep with $\rho=0.5$}} \\
0.00 & 0.50 & \best{62.83} & \best{62.54} & 97.80 & \second{58.33} & 94.76 & 91.69 & 34.38 & 25.42 & \second{98.56} & 15.39 & \best{0.504} & \best{0.443} & \second{1.428} \\
0.25 & 0.50 & 62.55 & 62.47 & 97.82 & 56.94 & 94.65 & 91.66 & 35.47 & \best{25.47} & 98.55 & \best{15.44} & \second{0.503} & \best{0.443} & \second{1.428} \\
0.50 & 0.50 & 62.64 & \second{62.50} & \best{97.84} & \best{59.72} & \best{94.94} & \best{92.10} & 35.47 & 25.41 & \best{98.57} & 15.38 & \second{0.503} & \second{0.444} & \second{1.428} \\
\rowcolor{rowgray} 0.75 & 0.50 & \second{62.70} & 62.46 & \second{97.83} & 55.56 & 94.84 & 91.97 & \second{36.70} & \second{25.44} & \best{98.57} & \second{15.42} & \second{0.503} & \best{0.443} & \best{1.427} \\
1.00 & 0.50 & 62.37 & 62.32 & 97.80 & 56.94 & \second{94.91} & \second{92.04} & \best{36.77} & 25.40 & \best{98.57} & 15.40 & \second{0.503} & \best{0.443} & \second{1.428} \\
\bottomrule
\end{tabular}}
\end{minipage}
\par\medskip

\subsection{Calibration-set sensitivity}
\label{app:calibration-prompts}

\par\medskip\noindent
\begin{minipage}{\linewidth}
\centering
\captionof{table}{Exact three-prompt calibration sets for the MiniMax-H3 composition study; prompts are reproduced verbatim.}
\label{tab:h3-calibration-prompt-sets}
\setlength{\tabcolsep}{6pt}
\renewcommand{\arraystretch}{1.12}
\small
\begin{tabularx}{\textwidth}{@{}>{\raggedright\arraybackslash}p{0.17\textwidth} >{\centering\arraybackslash}p{0.045\textwidth} X@{}}
\toprule
\rowcolor{rowgray}
\textbf{Calibration set} & \textbf{ID} & \textbf{Calibration prompt} \\
\midrule
\multirow[c]{3}{0.17\textwidth}{\raggedright\textbf{Complementary coverage}}
& \textcolor{pulseaccent}{\textbf{P1}} & A ceramic teapot rotating slowly on a wooden table in soft studio light \\
& \textcolor{pulseaccent}{\textbf{P2}} & A glass marble rolling across a linen cloth while the camera tracks beside it \\
& \textcolor{pulseaccent}{\textbf{P3}} & A small sailboat crossing a calm lake at dawn with gentle camera movement \\
\cmidrule(lr){1-3}
\multirow[c]{3}{0.17\textwidth}{\raggedright\textbf{Similar static}}
& \textcolor{pulseaccent}{\textbf{P1}} & A ceramic teapot resting on a wooden table under soft studio light, locked-off camera \\
& \textcolor{pulseaccent}{\textbf{P2}} & A polished red apple resting on a linen-covered table under soft studio light, locked-off camera \\
& \textcolor{pulseaccent}{\textbf{P3}} & A porcelain vase resting on a wooden shelf under soft studio light, locked-off camera \\
\cmidrule(lr){1-3}
\multirow[c]{3}{0.17\textwidth}{\raggedright\textbf{Heterogeneous}}
& \textcolor{pulseaccent}{\textbf{P1}} & A red fox trotting through a snowy pine forest while loose snow falls from the branches \\
& \textcolor{pulseaccent}{\textbf{P2}} & A vintage red train crossing a stone bridge in heavy rain with a tracking camera \\
& \textcolor{pulseaccent}{\textbf{P3}} & A hummingbird hovering beside red flowers in a sunlit rainforest, filmed with a macro lens \\
\cmidrule(lr){1-3}
\multirow[c]{3}{0.17\textwidth}{\raggedright\textbf{Similar dynamic}}
& \textcolor{pulseaccent}{\textbf{P1}} & A golden retriever walking from left to right through a green park while the camera tracks beside it \\
& \textcolor{pulseaccent}{\textbf{P2}} & A young woman walking from left to right along a quiet garden path while the camera tracks beside her \\
& \textcolor{pulseaccent}{\textbf{P3}} & A small delivery robot moving from left to right along a tree-lined walkway while the camera tracks beside it \\
\bottomrule
\end{tabularx}
\end{minipage}
\par\medskip

\par\medskip\noindent
\begin{minipage}{\linewidth}
\centering
\captionof{table}{MiniMax-H3 W4A4 calibration-set sensitivity on the VBench.}
\label{tab:h3-calibration-vbench326}
\setlength{\tabcolsep}{2.6pt}
\renewcommand{\arraystretch}{1.06}
\resizebox{\textwidth}{!}{%
\begin{tabular}{lc@{\hspace{6pt}}cccccccc@{\hspace{8pt}}ccc}
\toprule
\multirow[c]{2}{*}{Calibration set} & \multirow[c]{2}{*}{\# Prompts} & \multicolumn{8}{c}{VBench attributes (\%)} & \multicolumn{3}{c}{Dense-reference fidelity} \\
\cmidrule(lr){3-10}\cmidrule(lr){11-13}
 & & IQ $\uparrow$ & AQ $\uparrow$ & MS $\uparrow$ & DD $\uparrow$ & BC $\uparrow$ & SC $\uparrow$ & Scene $\uparrow$ & OC $\uparrow$ & PSNR $\uparrow$ & SSIM $\uparrow$ & LPIPS $\downarrow$ \\
\midrule
Dense BF16 & -- & 68.57 & 65.38 & 99.01 & 75.00 & 96.08 & 92.55 & 47.60 & 26.64 & ref. & ref. & ref. \\
\midrule
\multicolumn{13}{l}{\textit{Prompt count: shared ordered prefix}} \\
Prefix-1 & 1 & 66.39 & \second{65.55} & \second{98.73} & \second{68.06} & \best{95.33} & 92.42 & \second{46.51} & 25.97 & 14.26 & 0.424 & 0.509 \\
Prefix-3 & 3 & \best{66.83} & 65.43 & \second{98.73} & \best{70.83} & 94.97 & 92.23 & \best{46.88} & \best{26.24} & 14.17 & 0.424 & 0.514 \\
Prefix-6 & 6 & \second{66.74} & \best{65.75} & \best{98.74} & \second{68.06} & \second{95.29} & \best{92.65} & 40.84 & 26.15 & 14.22 & 0.426 & 0.512 \\
Prefix-10 & 10 & 65.87 & 65.37 & \best{98.74} & \best{70.83} & 95.12 & 92.42 & 44.62 & 26.11 & \best{14.40} & \best{0.434} & \best{0.501} \\
Prefix-20 & 20 & 65.52 & 65.18 & 98.70 & \second{68.06} & 95.16 & \second{92.47} & 43.60 & \second{26.21} & \second{14.31} & \second{0.431} & \second{0.506} \\
\midrule
\multicolumn{13}{l}{\textit{Prompt composition: fixed three-prompt budget}} \\
Complementary coverage & 3 & \second{66.83} & 65.43 & \second{98.73} & \best{70.83} & 94.97 & 92.23 & \best{46.88} & \best{26.24} & 14.17 & \second{0.424} & 0.514 \\
Similar static & 3 & 65.30 & \second{65.64} & \best{98.79} & 65.28 & 95.24 & 92.17 & 42.30 & \second{26.21} & \best{14.39} & \best{0.430} & \second{0.508} \\
Heterogeneous & 3 & \best{67.52} & \best{66.18} & 98.66 & 65.28 & \best{95.45} & \best{92.77} & 44.04 & \second{26.21} & \second{14.22} & 0.422 & \best{0.507} \\
Similar dynamic & 3 & 66.49 & 65.61 & 98.67 & \second{66.67} & \second{95.26} & \second{92.67} & \second{44.48} & 26.08 & 14.12 & 0.421 & 0.511 \\
\bottomrule
\end{tabular}}
\end{minipage}
\par\medskip

We separately study calibration-set size and prompt composition on MiniMax-H3 W4A4 in Table~\ref{tab:h3-calibration-vbench326}. Dense BF16 is included as a common reference and is not part of either sweep.

\paragraph{Prompt count.}
The count sweep uses nested prefixes of one ordered pool, so larger sets retain all earlier prompts. Increasing the calibration set does not uniformly improve quality: ten prompts give the best paired fidelity, whereas three prompts achieve the highest Overall Consistency and Scene scores in this sweep. The twenty-prompt set does not improve on ten prompts in any paired fidelity metric, indicating that a larger calibration budget alone does not ensure better reconstruction.

\paragraph{Prompt composition.}
At a fixed three-prompt budget, composition changes the balance between appearance, motion, and fidelity. Heterogeneous prompts achieve the best image and aesthetic quality and LPIPS, while complementary coverage leads Dynamic Degree, Scene, and Overall Consistency. Similar-static prompts yield the best PSNR, SSIM, and Motion Smoothness. We use complementary coverage to span object appearance, object motion, and camera motion, with strong scene and overall consistency in this comparison. Prefix-3 and complementary coverage use the same checkpoint and are repeated to make the two sweeps self-contained. Table~\ref{tab:h3-calibration-prompt-sets} lists the exact prompts.

\subsection{Propagation-risk objective}
\label{app:risk-hyperparameters}
Table~\ref{tab:wan13-hparam-sensitivity} varies the tail fraction $\rho$ at fixed $\lambda=0.75$ and the tail weight $\lambda$ at fixed $\rho=0.5$. In the $\rho$ sweep, the smallest fraction gives the best paired fidelity, while $\rho=0.5$ yields the highest Overall Consistency. The $\lambda$ sweep shows limited variation in paired fidelity, with PSNR between 15.38 and 15.44~dB and LPIPS between 0.443 and 0.444. Other attributes favor different settings, so no configuration dominates all metrics. We retain $(\lambda,\rho)=(0.75,0.5)$ as a balanced default; it also gives the lowest reported normalized jerk in both sweeps.

\section{Native W4A4 deployment and profiling}
\label{app:deployment-details}
This section provides the implementation details underlying the efficiency results in Section~\ref{sec:efficiency}. We describe the packed linear path, the applied fusions, and the measurement boundary used for native W4A4 inference.

\subsection{Packed linear path}
\label{app:packed-linear}
We deploy Wan~2.1-1.3B on a single RTX~5080 (SM120) using PyTorch~2.11 and CUDA~12.8. Corrected weights and scales are exported offline and stored in packed NVFP4 format, while activations are quantized dynamically during inference. Among the 300 packed linear modules, 210 supported modules dispatch to native W4A4 Tensor Core kernels. The remaining operators run at the original model precision and are included in the reported DiT-step latency.

\subsection{Fusion and workspace reuse}
\label{app:fusion-workspace}
We fuse the RPBH sign transform with activation packing, fuse the QKV projections, and reuse stream-ordered activation workspaces and invariant BF16 masks across calls. These optimizations avoid repeated allocation, host synchronization, and explicit materialization of the rotated activations. For a representative FFN shape of $(M,K,N)=(32760,1536,8960)$, the fused path takes 1.54~ms, compared with 7.74~ms for BF16 GEMM, corresponding to a 5.01$\times$ speedup. It is also 1.18$\times$ faster than the path that materializes the rotation. Across six sequences, workspace reuse produces zero maximum deviation from freshly allocated workspaces.

\subsection{Attention backend and measurement boundary}
\label{app:attention-measurement}
SageAttention2 uses an SM120 CUDA kernel with INT8 $QK^\top$ and FP8 $PV$ computation, while VC Attention additionally fuses V-Smooth into this path. The persistent workspaces add approximately 185~MiB. Before workspace reuse, the combined VC and SageAttention2 paths require 4.07 and 3.69~s per DiT step, respectively. Removing allocation and materialization overhead yields a 2.12$\times$ backend speedup. We synchronize all measurements and report single-video DiT-step latency and DiT-only peak allocation. Offline calibration and packing, text encoding, and VAE decoding are excluded from both measurements.

\section{Additional qualitative results}
\label{app:qualitative}
This section provides additional visual comparisons under controlled generation settings. All panels use matched prompts, seeds, schedulers, and initial noise. Red labels report full-video PSNR and SSIM against the corresponding BF16 trajectory. We sample frames at fixed temporal positions to reveal both appearance drift and long-range temporal inconsistency.

\subsection{Wan 2.1-1.3B W4A4 comparisons}
\label{app:qual-wan13-w4a4}
\begin{figure*}[t]
  \centering
  \includegraphics[width=\textwidth]{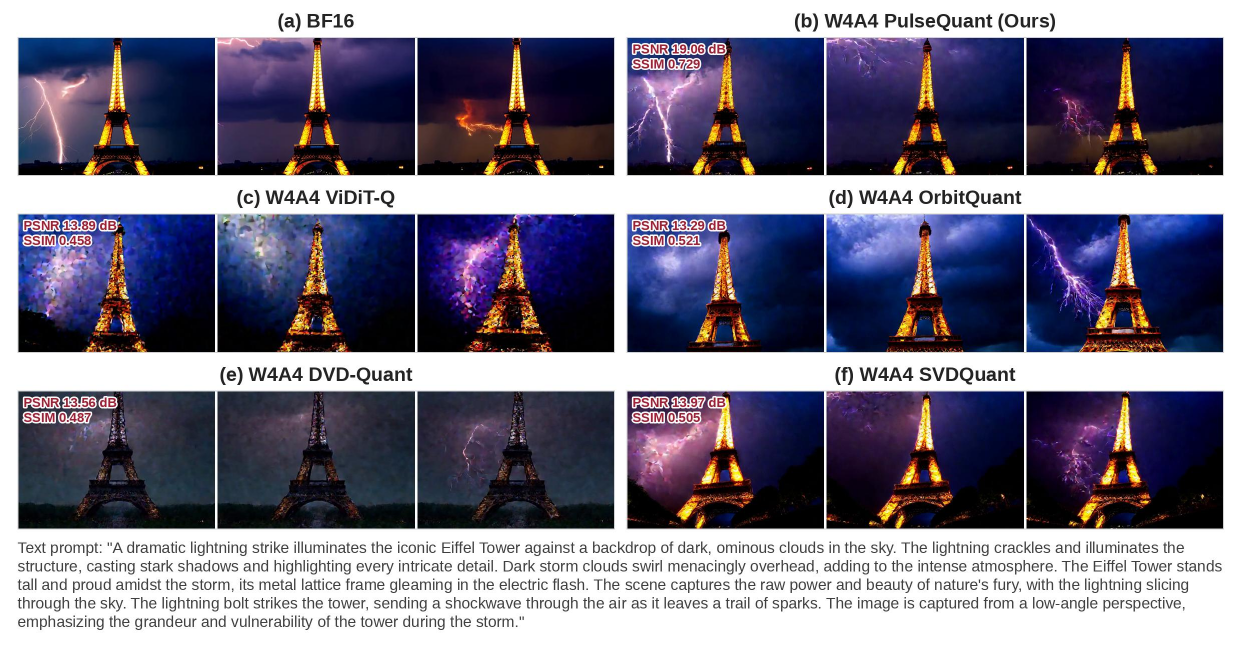}

  \caption{Wan~2.1-1.3B W4A4 comparison for the Eiffel Tower under lightning.}

  \label{fig:wan13-qual-223}
\end{figure*}

\begin{figure*}[t]
  \centering

  \includegraphics[width=\textwidth]{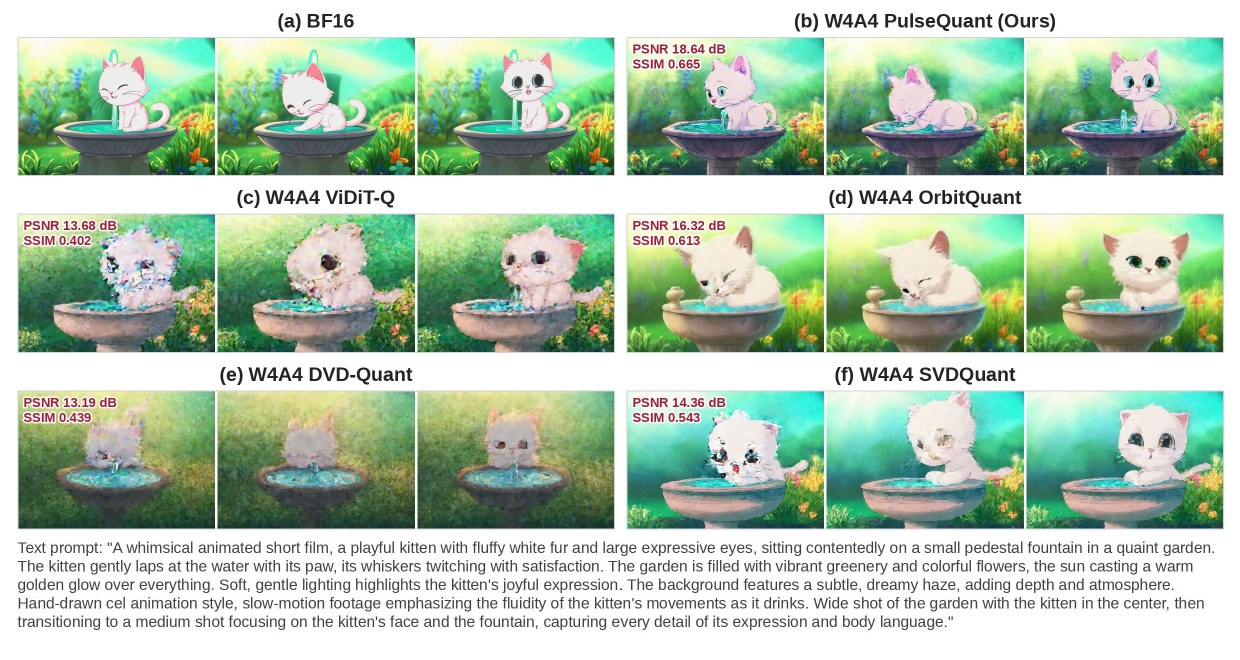}

  \caption{Wan~2.1-1.3B W4A4 comparison for an animated kitten at a garden fountain.}
  \label{fig:wan13-qual-118}
\end{figure*}

\begin{figure*}[t]
  \centering

  \includegraphics[width=\textwidth]{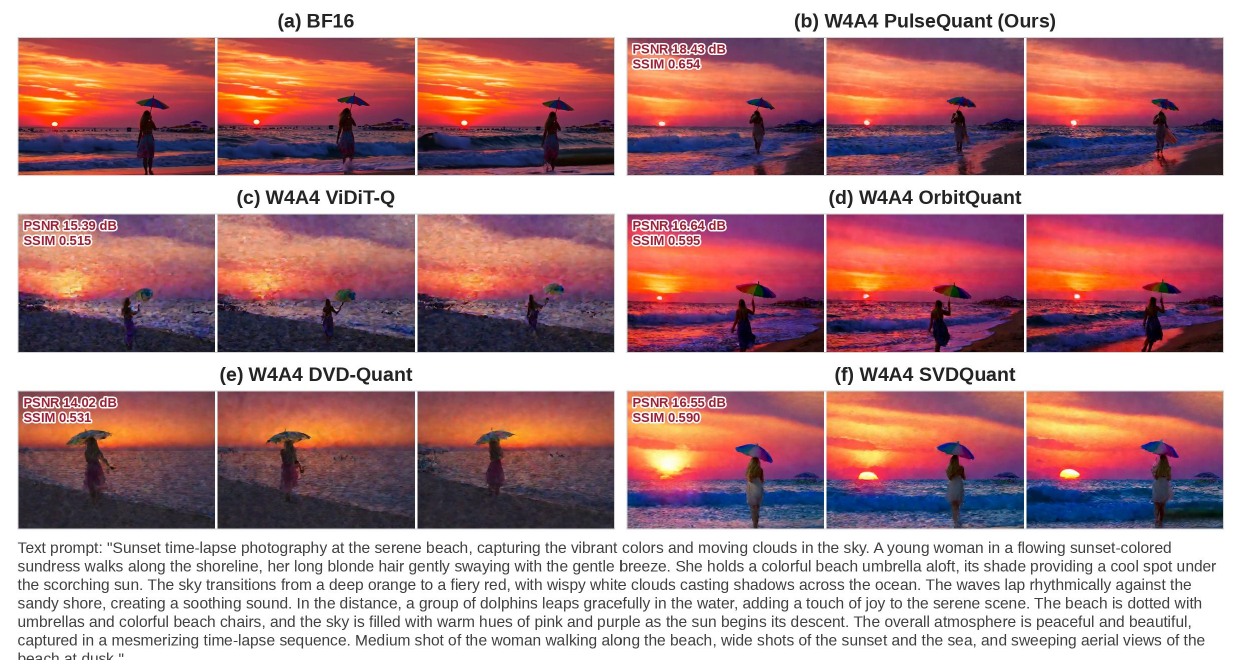}

  \caption{Wan~2.1-1.3B W4A4 comparison for a woman walking along a beach at sunset.}
  \label{fig:wan13-qual-153}
\end{figure*}

Figures~\ref{fig:wan13-qual-223}, \ref{fig:wan13-qual-118}, and~\ref{fig:wan13-qual-153} compare six methods on three fixed prompts using frames 5, 40, and 75. We select the cases once according to the PulseQuant--ViDiT-Q paired-fidelity gap and then filter them for semantic diversity. None of these prompts is used for calibration.

The kitten sequence tests fine foreground geometry and foreground--background separation, whereas the lightning and beach scenes stress global composition and illumination. Across distant frames, PulseQuant better preserves the principal subject and scene layout. The competing methods more frequently alter silhouette, texture, contrast, or background structure. These errors persist across time, showing how trajectory-level quantization error develops into semantic drift.

\subsection{W4A6 and cross-architecture comparisons}
\label{app:qual-transfer}
\begin{figure*}[t]
  \centering
  \includegraphics[width=\textwidth]{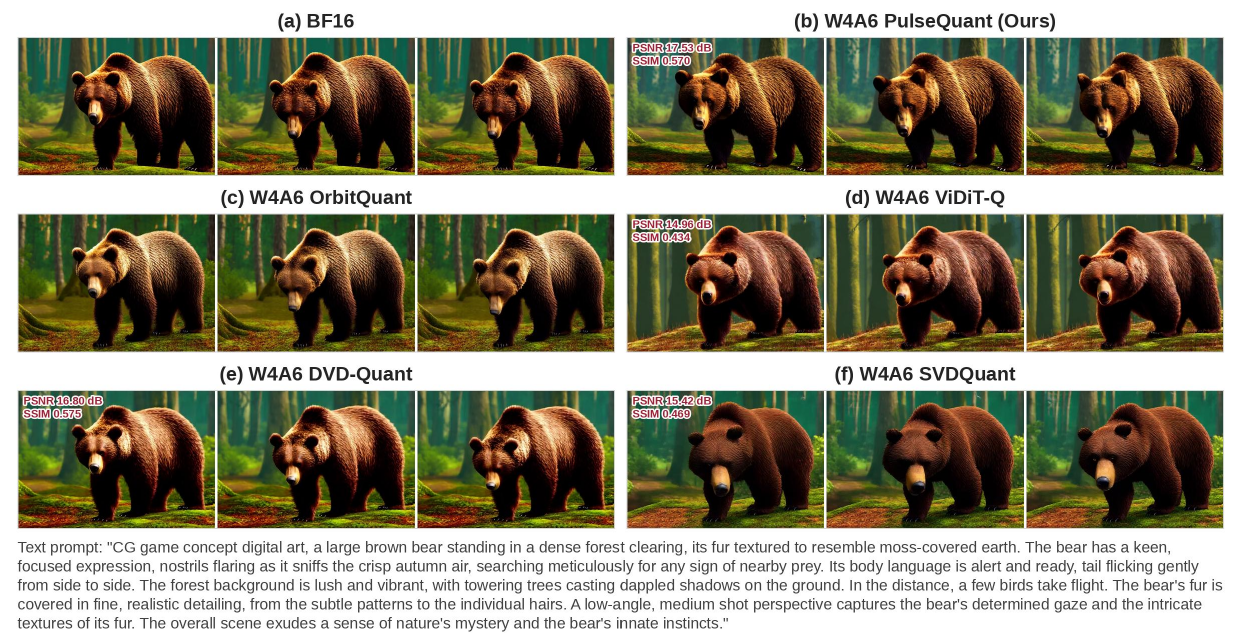}
  \caption{Wan~2.1-1.3B W4A6 comparison for a brown bear in a forest clearing.}
  \label{fig:wan13-qual-w4a6-138}
\end{figure*}

\begin{figure*}[t]
  \centering
  \includegraphics[width=\textwidth]{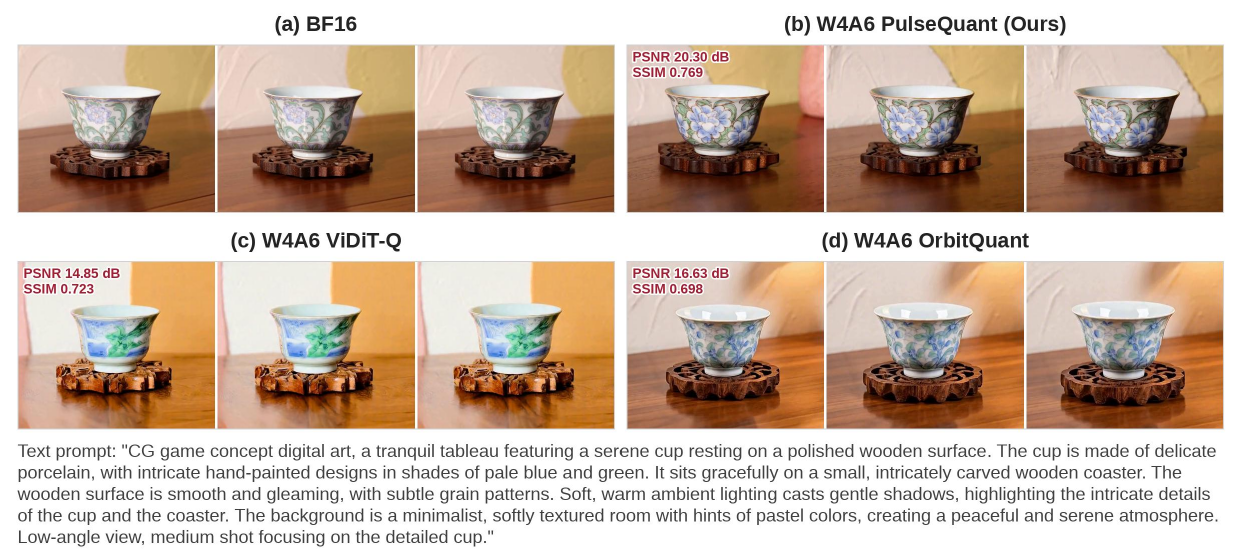}
  \caption{MiniMax-H3 W4A6 comparison for a porcelain cup on a carved wooden coaster.}
  \label{fig:h3-qual-w4a6-25}
\end{figure*}

\begin{figure*}[t]
  \centering
  \includegraphics[width=\textwidth]{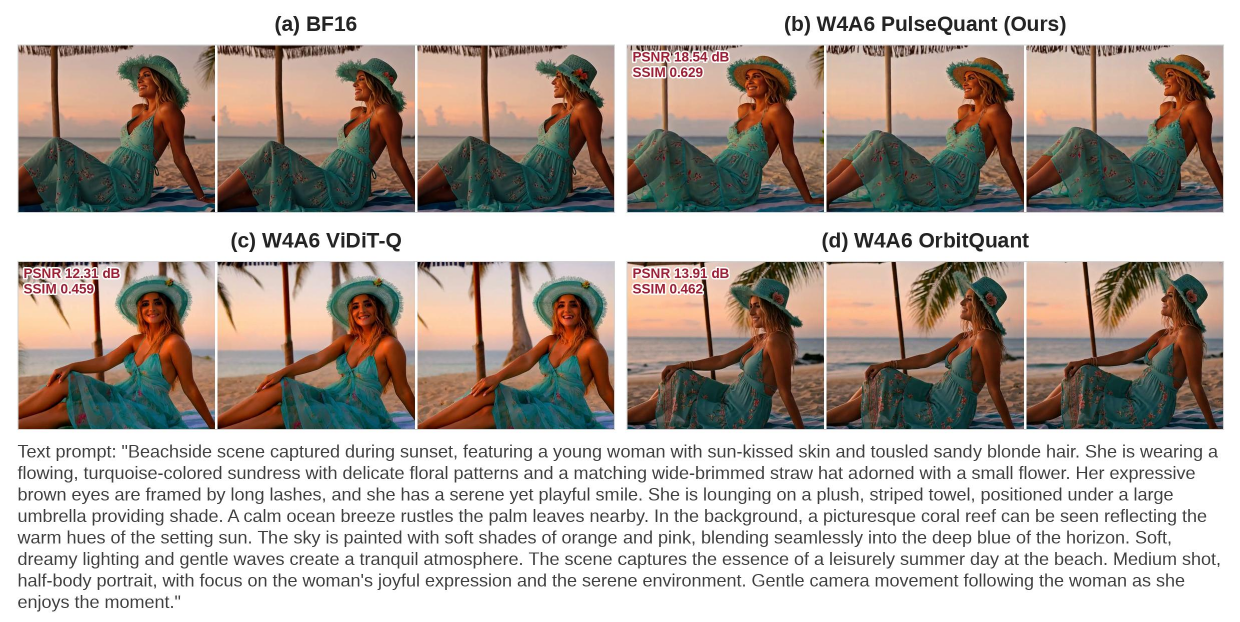}
  \caption{MiniMax-H3 W4A6 comparison for a woman relaxing by the beach at sunset.}
  \label{fig:h3-qual-w4a6-251}
\end{figure*}

\begin{figure*}[t]
  \centering
  \includegraphics[width=\textwidth]{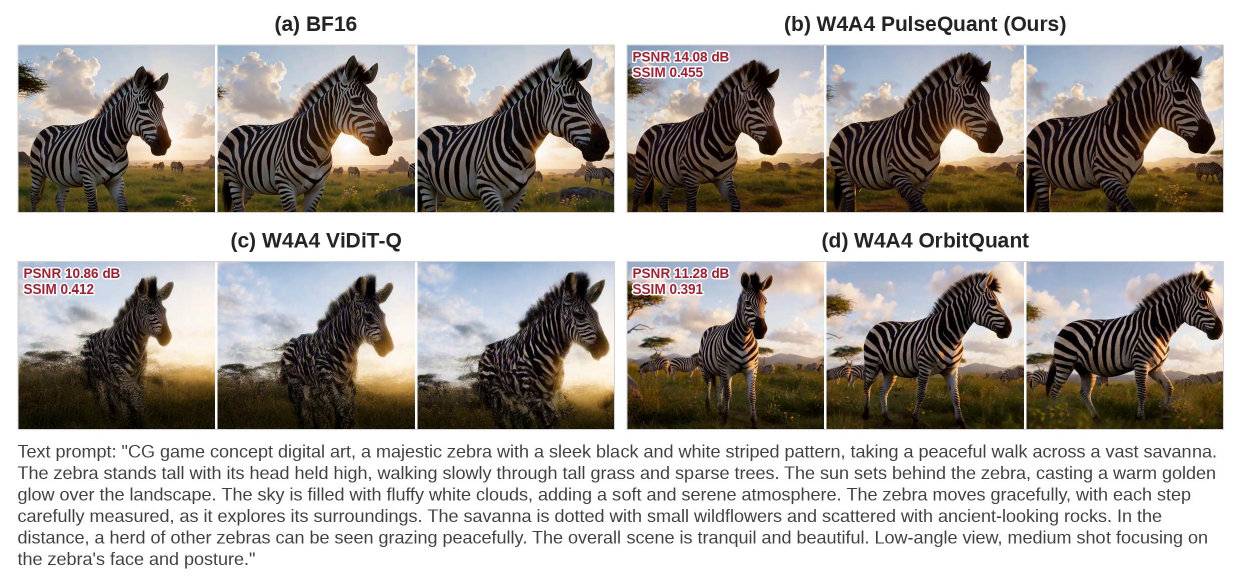}
  \caption{MiniMax-H3 W4A4 comparison for a zebra walking across a sunlit savanna.}
  \label{fig:h3-qual-w4a4-143}
\end{figure*}

\begin{figure*}[t]
  \centering
  \includegraphics[width=\textwidth]{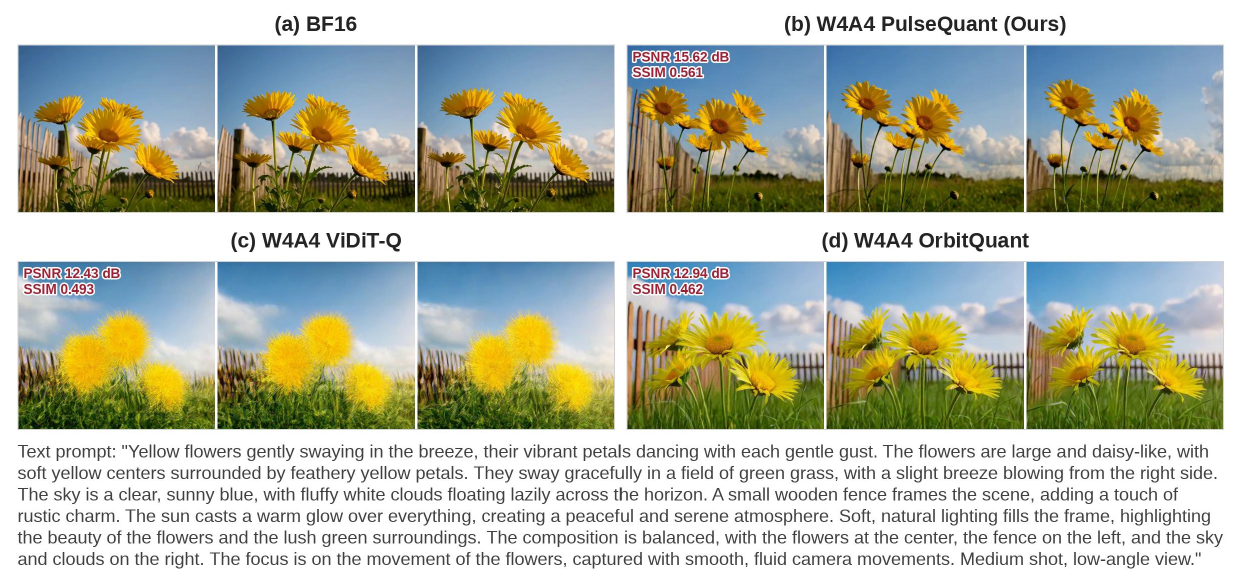}
  \caption{MiniMax-H3 W4A4 comparison for yellow flowers swaying beside a wooden fence.}
  \label{fig:h3-qual-w4a4-239}
\end{figure*}

Figure~\ref{fig:wan13-qual-w4a6-138} presents a six-method comparison on Wan~2.1-1.3B W4A6. Figures~\ref{fig:h3-qual-w4a6-25}--\ref{fig:h3-qual-w4a4-239} apply the same matched protocol to MiniMax-H3 at W4A6 and W4A4. We show frames 5, 40, and 75 for Wan~2.1-1.3B and frames 5, 60, and 100 for MiniMax-H3. The examples cover rigid fine structure, human appearance, articulated animal motion, and repeated floral texture. Across these settings, PulseQuant remains closer to the dense subject geometry and composition, whereas competing outputs more often alter texture, scale, or pose.

\clearpage

\end{document}